\documentclass[runningheads]{llncs}
 
\usepackage{accv}

\usepackage{accvabbrv}
\usepackage[table]{xcolor}
\usepackage{graphicx}
\usepackage{booktabs} 
\usepackage{amssymb}
\usepackage{multirow}
\usepackage{subcaption}

\usepackage[accsupp]{axessibility}  

\usepackage[breaklinks,colorlinks,citecolor=accvblue]{hyperref}
\usepackage{orcidlink}
\usepackage{longtable}
\begin{document}

\title{TIRAD: Text-guided Image Residual Learning for Generalist Anomaly Detection} 

\titlerunning{TIRAD}

\author{Seongsu Lee\orcidlink{0009-0005-0098-3179} \and
Seunghyun Oh\orcidlink{0009-0007-4660-5725} \and
Seungjae Moon\orcidlink{0009-0007-3476-1878} \and
Youngmin Ro\textsuperscript{*}\orcidlink{0000-0002-0661-9192}}

\authorrunning{Lee et al.}

\institute{Machine Intelligence Laboratory, \\ Dept. of Artificial Intelligence, University of Seoul, Korea\\
\email{\{ssulee0206, osh1795, msj0243, youngmin.ro\}@uos.ac.kr}\
}

\maketitle
\begingroup
\renewcommand{\thefootnote}{*}
\footnotetext{Corresponding author.}
\endgroup

\begin{abstract}
    Recent anomaly detection studies have shown that residual features, which model the discrepancy between normal and abnormal patterns, provide strong generalization to anomalies unseen during training. However, existing residual-based approaches mainly rely on visual discrepancies derived from image representations, while semantic discrepancies that can be explicitly expressed through language remain largely underexplored. To address this problem, we propose TIRAD, Text-guided Image Residual Learning for Generalist Anomaly Detection, a multimodal residual learning framework that defines the discrepancy between textual representations of normal and abnormal states as a Text Residual Feature and uses it to guide the Image Residual Feature. Specifically, a Text Fusion Module converts patch-level text residuals into a gate that modulates the image residual of each patch. The gate retains the visual discrepancies that the text residual identifies as defect-related and attenuates the others, leading to more discriminative anomaly representations. Extensive experiments on industrial benchmarks and industrial-to-medical transfer demonstrate that the proposed method outperforms existing image-centric residual learning methods.

\end{abstract}


\section{Introduction}
\label{sec:intro}

Anomaly Detection (AD)~\cite{padim, spade, patchcore, winclip} aims to identify abnormal patterns that deviate from normal observations and has been widely adopted in industrial inspection and medical diagnosis~\cite{mediclip,mvfa, medical_anomalydino, madclip}. Since anomalous samples are inherently rare, diverse, and often unpredictable, collecting sufficient defect data for every possible abnormal scenario is impractical. This difficulty becomes more severe in real-world deployment, where new object categories, unseen defect types, and domain shifts continuously appear. Consequently, recent studies have increasingly focused on Generalist Anomaly Detection (GAD)~\cite{inctrl, resad, nagl}, which seeks to develop a unified model~\cite{you2022unified, uni_uninet} capable of detecting previously unseen anomalies across diverse object categories and domains without category-specific retraining. Unlike conventional AD settings that mainly assume the same category distribution between training and testing, GAD requires a model trained on known categories to generalize to unseen target categories and unseen anomaly types.

\begin{figure}[t]
    \centering
    \includegraphics[width=1 \linewidth,trim={10pt 175pt 10pt 200pt}, clip]{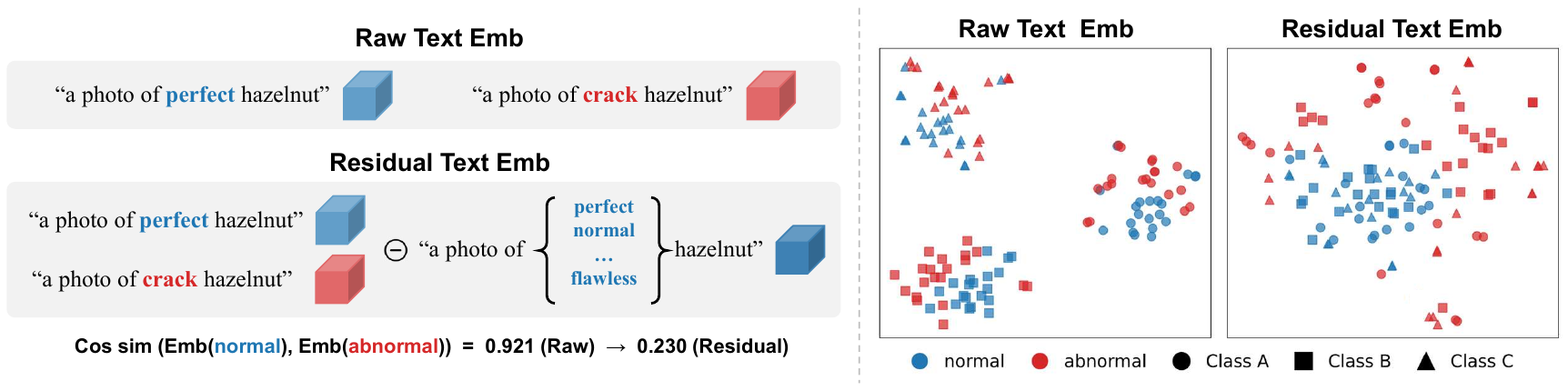}
    \caption{(a) Illustration of the text residual: subtracting the normal text embedding from the abnormal one yields a text residual that markedly reduces the normal--abnormal similarity. (b) Comparison of text features: raw normal and abnormal embeddings are highly similar and entangled, whereas the text residual features separate normal from abnormal consistently across object classes in the t-SNE projection.}
    \label{fig:GAD}
\end{figure}

A key challenge in GAD is to learn anomaly-relevant representations that are less dependent on object-specific appearance. Recent studies address this issue from the perspective of residual learning. In anomaly detection, a residual can be interpreted as the discrepancy between a normal state and an abnormal state. Instead of representing an object by its absolute visual appearance, residual features emphasize how the query deviates from normality. By suppressing category-specific characteristics, residual representations provide transferable cues for detecting unseen anomalies.

Building on this idea, representative residual-based methods~\cite{inctrl, resad, yao2026adpretrain, nagl} exploit image residual features by comparing query image features with normal reference features or by modeling normal--abnormal visual discrepancies, and demonstrate strong generalization to unseen categories. This line of research indicates that residual learning is a promising direction for building generalist anomaly detectors.

However, existing residual-based anomaly detection methods define residuals only in the image feature space. In other words, they assume that abnormality can be sufficiently represented by visual discrepancies between normal and abnormal image features. Although image residuals are effective for many appearance-level defects, they respond to any deviation from the reference features, including variations in texture, pose, or imaging conditions that are not defects. Whether a visual discrepancy is actually anomalous is a semantic judgment, and the defect concepts needed for this judgment, such as a crack, a scratch, or a missing part, can be explicitly described through language. Relying only on image residuals therefore leaves the model without a semantic criterion for deciding which discrepancies correspond to true anomalies, particularly when it must generalize to unseen categories and domains.

Meanwhile, several AD methods based on Vision-Language Models (VLMs) have attempted to incorporate textual information through handcrafted prompts ~\cite{winclip,sadikaj2025multiads} or learnable text representations~\cite{anomalyclip, promptad, zhou2022CoOp}. However, these methods typically use the resulting text embeddings as semantic references and compute anomaly scores through image-text similarity~\cite{adaclip, gu2024anomalygpt}. Such direct reliance on text embeddings is suboptimal for anomaly detection because VLMs are primarily trained for category-level image-text alignment rather than fine-grained discrimination between normal and abnormal states. As a result, normal and abnormal textual descriptions may remain entangled in the original text feature space, limiting their ability to provide reliable anomaly-aware cues. More importantly, although residual learning has been extensively studied in the image feature space, textual residuals remain largely underexplored in anomaly detection.

To address this gap, we introduce the Text Residual Feature. Given a set of normal and abnormal text candidates, each query patch is softly matched to the candidates, and a normal text anchor is subtracted from the matched text feature. The resulting residual is close to zero for patches consistent with normality and otherwise points toward the defect semantics most consistent with the patch. Removing the shared normal component leaves the part of the text feature that differs from normality. As illustrated in Fig.~\ref{fig:GAD}(a), this subtraction sharply reduces the similarity between normal and abnormal descriptions (cosine similarity $0.921 \rightarrow 0.230$). Since defect concepts are often shared across different object categories and domains, text residuals are more transferable for generalist anomaly detection. As further shown in Fig.~\ref{fig:GAD}(b), while raw normal and abnormal text features remain entangled, text residual features separate the two consistently across object classes in the t-SNE projection.

Based on this idea, we propose \textbf{TIRAD} (\textbf{T}ext-guided \textbf{I}mage \textbf{R}esidual learning for generalist \textbf{A}nomaly \textbf{D}etection), a multimodal residual learning framework in which text serves as a semantic refinement of image residuals. The image branch computes patch-level image residuals by measuring the discrepancy between query image features and normal reference features, following a nearest-neighbor residual formulation. In parallel, the text branch computes a text residual for each patch by soft matching over multiple normal and abnormal candidates. This text residual is converted into a scalar gate that multiplicatively modulates the image residual of the same patch, producing a text-guided residual score that is merged with the image-only residual score for the final prediction. Because the gate can only attenuate or preserve the image residual, the text branch does not localize anomalies on its own; it retains the visual discrepancies that the text residual identifies as defect-related and attenuates the others, without any per-category parameter.

The main contributions of this work are summarized as follows:
\begin{itemize}
    \item We introduce the Text Residual Feature, a semantic residual representation that explicitly models the discrepancy between normal and abnormal textual features, addressing the largely underexplored use of residual cues in the textual domain for generalist anomaly detection.

    \item We propose TIRAD, a multimodal residual learning framework that leverages text residual features to guide image residual features. Through a Text Fusion Module, our method effectively combines semantic transferability with visual sensitivity.

    \item Extensive experiments on diverse anomaly detection benchmarks show that our method improves generalization to unseen categories and anomaly types including a challenging cross-domain, industrial-to-medical transfer setting and outperforms existing image-centric residual learning approaches.
\end{itemize}

\section{Related Work}
\label{sec:related}
\noindent\textbf{Generalist and Residual-based Anomaly Detection. }
Classical anomaly detection assumes abundant normal data and a shared category distribution between training and testing, modeling normal patterns through memory banks, density estimation, reconstruction-based networks, or normalizing flows~\cite{patchcore, padim, spade, reconpatch, zavrtanik2021draem, deng2022anomaly, you2022unified, yu2021fastflow, gudovskiy2022cflow}. Few-shot methods reduce this dependence but often require retraining on the target category and transfer poorly to new domains~\cite{regad, graphcore, li2024onetonormal, IIPAD, gao2025learning}, motivating the generalist setting, where a single model trained on known categories is applied directly to unseen categories without retraining~\cite{inctrl, adaptclip}. A promising direction toward this goal is residual learning, where each query patch is matched to its nearest normal reference and the difference is taken as a residual feature; by cancelling category-specific appearance, residuals stay small for normal regions and grow for anomalies, providing class-invariant cues~\cite{resad, zhang2023prototypical}. Subsequent methods further exploit normal--abnormal reference residuals to learn transferable abnormal patterns~\cite{nagl, yao2026adpretrain}. 

\noindent\textbf{Vision-Language Models for Anomaly Detection. }
Vision-language models such as CLIP~\cite{clip} have been widely adopted for anomaly detection through textual descriptions of normal and abnormal states. WinCLIP applies handcrafted prompt ensembles for zero-/few-shot detection~\cite{winclip, li2024musc, chen2023VAND}, while later work learns object-agnostic or normal-only prompts to better separate normal and abnormal semantics~\cite{zhou2022CoOp, Zhou_2022_CVPR_CoCoOp, anomalyclip, promptad, adaclip}. Beyond prompt design, recent methods incorporate large vision-language models or defect-aware supervision for finer anomaly understanding~\cite{gu2024anomalygpt, sadikaj2025multiads, oh2025aulora}, and some approaches even revisit vision-only foundation features for language-free detection~\cite{hou2026visualad, uni_uninet}. However, these methods use the resulting text embeddings as static semantic anchors and score anomalies through image--text similarity. Because vision-language models are pre-trained for category-level image--text alignment rather than fine-grained normal/abnormal discrimination, normal and abnormal descriptions often remain entangled in the original text space. Consequently, although residual learning has been extensively studied for image features, computing residuals within the text feature space remains largely underexplored. Our work explores this direction by using patch-level text residuals to guide image residuals, combining the transferability of semantic residuals with the localization ability of visual matching.

\section{Method}
\label{sec:method}

\begin{figure}[t]
    \centering
    \includegraphics[width=1.0\linewidth,trim={50pt 110pt 50pt 120pt}, clip]{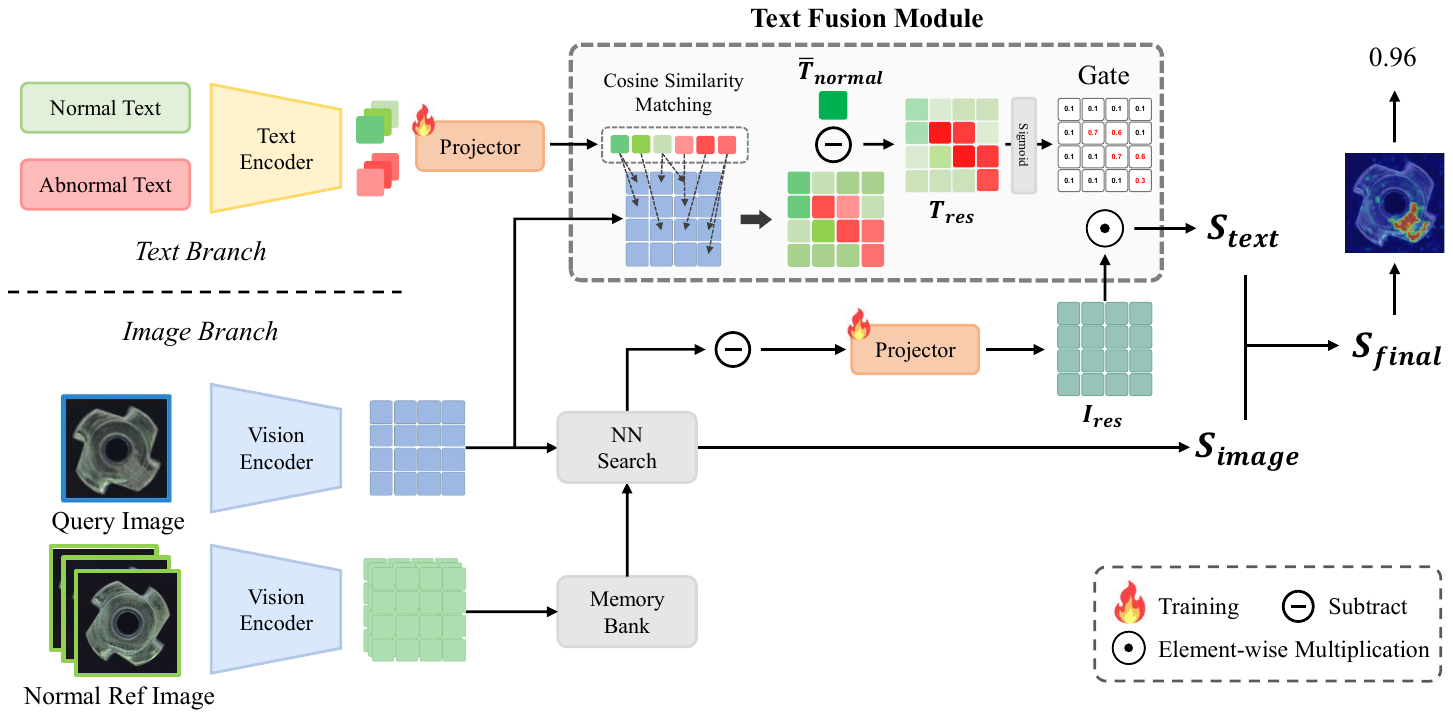}
    \caption{Overview of the proposed framework. The image branch uses memory-bank-based nearest-neighbor search to compute an image anomaly score map $S_{\mathrm{image}}$ from query and normal reference features. In parallel, the text branch constructs text residual features from normal and abnormal prompts, and the Text Fusion Module converts them into a gate that semantically modulates image residual features to produce a text-guided score map $S_{\mathrm{text}}$. The two score maps are merged into the final anomaly score map $S_{\mathrm{final}}$, which is used for both image-level anomaly detection and pixel-level anomaly localization.
    }
    \label{fig:main_figure}
\end{figure}

We propose TIRAD, a text residual-based framework for generalist anomaly detection.
Given a query image and a few normal reference images from the target category, our goal is to produce both an image-level anomaly score and a pixel-level anomaly localization map without updating the frozen image backbone at test time.
The key idea is to complement conventional image residuals with \emph{text residuals}, which describe patch-wise semantic deviations from the normal textual state.
While image residuals capture visual discrepancies between the query and normal references, text residuals provide defect-aware semantic guidance that suppresses visually inconsistent but semantically normal variations.

As illustrated in Fig.~\ref{fig:main_figure}, our framework comprises two complementary branches.
The image branch (Sec.~\ref{sec:image_branch}) produces a training-free anomaly map $S_{\mathrm{image}}$ from a normal reference memory bank, while the text branch (Sec.~\ref{sec:text_branch}) derives a text residual that the Text Fusion Module converts into a gate, modulating the image residual to yield a text-guided map $S_{\mathrm{text}}$.
The two maps are merged by a learnable weight (Sec.~\ref{sec:fusion}) into the final anomaly map $S_{\mathrm{final}}$ for both image-level detection and pixel-level localization.

\subsection{Problem Formulation}
\label{sec:formulation}

We consider a generalist anomaly detection setting where a model is trained on known source categories and evaluated on unseen target categories.
Let $\mathcal{D}_{\mathrm{src}}=\{(x_i, y_i, m_i)\}_{i=1}^{N}$ denote the source training set, where $x_i$ is an image, $y_i\in\{0,1\}$ is the image-level anomaly label, and $m_i\in\{0,1\}^{H\times W}$ is the pixel-level anomaly mask.
Normal images have $y_i=0$ and an all-zero mask. During training, each episode consists of a query image $x^q$, its label $y^q$, mask $m^q$, and a set of $K$ normal reference images $\mathcal{R}^{n}=\{x^{n}_{k}\}_{k=1}^{K}$ sampled from the source normal images.

At test time, the model is evaluated on unseen target categories that are not used during training.
For each target query image, only a few normal reference images from the same target category are provided.
No target-category fine-tuning is performed.
The objective is to learn a category-generalizable anomaly scoring function $\mathcal{M}(x^q,\mathcal{R}^{n},c) \rightarrow (S, s)$, where $c$ denotes the category name, $S\in[0,1]^{H\times W}$ is the anomaly localization map, and $s\in[0,1]$ is the image-level anomaly score.

\subsection{Image Branch}
\label{sec:image_branch}

The image branch aims to capture patch-level visual discrepancies between a query image and a few normal reference images. Given a query image $x^q$ and $K$ normal reference images $\{x^n_k\}_{k=1}^{K}$, we first extract patch-level visual features using a frozen vision encoder $\phi(\cdot)$~\cite{dinov2, dinotxt} as $F^q=\phi(x^q)=\{f^q_i\}_{i=1}^{L}$ with $f^q_i\in\mathbb{R}^{C}$, where $L=h\times w$ denotes the number of image patches. The same encoder is applied to each normal reference image, and all reference patch features are collected to construct a normal memory bank~\cite{patchcore, resad} $\mathcal{M}^{n}=\bigcup_{k=1}^{K}\phi(x^n_k)$.

For each query patch feature $f^q_i$, we retrieve its nearest normal reference feature from the memory bank by cosine similarity, $\hat{f}^{n}_i = \arg\max_{f \in \mathcal{M}^{n}} \mathrm{sim}(f^q_i,f)$, where $\mathrm{sim}(\cdot,\cdot)$ denotes cosine similarity. The retrieved feature $\hat{f}^{n}_i$ represents the closest normal appearance corresponding to the query patch. We then define the image residual feature as the difference between the query patch feature and its nearest normal reference feature, $r^I_i = f^q_i - \hat{f}^{n}_i$. We further apply a lightweight image projector $g_I(\cdot)$ to obtain the projected image residual $\tilde{r}^{I}_i = g_I(r^I_i)$, which is later modulated by the text-derived gate in the text branch.

The image residual feature describes how the local visual pattern of the query image deviates from normal reference appearances. For normal regions, the query patch is expected to be close to at least one normal reference patch, leading to a small residual. In contrast, anomalous regions are likely to produce larger residuals because their visual patterns cannot be well matched by the normal memory bank.

We also compute an image score map $S_{\mathrm{image}}$ using the cosine distance between each query patch and its nearest normal reference patch, $S_{\mathrm{image}, i} = 1 - \mathrm{sim}(f^q_i, \hat{f}^{n}_i)$ for $i=1,\dots,L$, which yields the image score map $S_{\mathrm{image}}=\{S_{\mathrm{image}, i}\}_{i=1}^{L}$. The image branch therefore provides two outputs: the projected image residual features $\{\tilde{r}^{I}_i\}_{i=1}^{L}$, which are later modulated by the text branch  via the Text Fusion Module, and the score map $S_{\mathrm{image}}$, which is used in the final merge. Since this prior is based solely on visual matching, we further introduce a text branch to incorporate semantic residual cues into the scoring process.

\subsection{Text Branch}
\label{sec:text_branch}

The text branch provides semantic guidance for image residual features. While the image branch measures visual deviations from normal reference images, the text branch estimates whether each query patch is semantically closer to a normal state or an abnormal state. To this end, it derives a patch-wise text residual and, through the Text Fusion Module, turns this residual into a gate that modulates the image residual and produces a text-guided score; the following paragraphs detail each component.

Given the category name, we construct two sets of predefined text prompts: normal-state prompts and abnormal-state prompts. The normal prompts describe defect-free conditions of the object, such as \textit{perfect} and \textit{flawless}. The abnormal prompts describe defect types such as \textit{scratch}, \textit{crack}, \textit{broken}, \textit{missing}, and other anomaly-related states. We denote the resulting normal and abnormal text features as $T^n=\{t^n_p\}_{p=1}^{P}$ and $T^a=\{t^a_m\}_{m=1}^{M}$, respectively.

Since frozen text embeddings are not directly optimized for patch-level anom\-a\-ly discrimination, we use a lightweight text projector $g_T(\cdot)$ to adapt them to the residual-based anomaly detection objective, yielding $\tilde{t}^{n}_p = g_T(t^n_p)$ and $\tilde{t}^{a}_m = g_T(t^a_m)$. We obtain a normal text anchor by aggregating the projected normal text features, $t^n_{\mathrm{anc}} = \mathrm{Norm}\!\left(\frac{1}{P}\sum_{p=1}^{P}\tilde{t}^{n}_p\right)$. The projected normal and abnormal text features are then used as semantic candidates for patch-wise text matching.

For each query patch feature $f^q_i$, we compare it with the projected text candidates $\tilde{t}_j \in \tilde{T}^n \cup \tilde{T}^a$ and softly match each patch to the most relevant textual state. The matching weights are obtained by applying a temperature-scaled softmax, with temperature $\tau$, over the cosine similarities between the patch feature and all projected candidates. The matched text representation $t^{\mathrm{match}}_i$ for the $i$-th patch is then computed as the weighted average of the projected candidates under these matching weights.

We define the text residual feature as the discrepancy between the matched text representation and the normal text anchor:
\begin{equation}
    r^T_i = t^{\mathrm{match}}_i - t^n_{\mathrm{anc}}.
\end{equation}
This residual represents how far the semantic state of a patch deviates from the normal textual state. If a patch is normal, its matched text representation is expected to be close to the normal anchor, resulting in a small text residual. If a patch contains a defect, the matched representation tends to move toward abnormal text candidates, producing a larger semantic residual.

The text residual feature is then used to guide the image residual feature. Specifically, we map the text residual through a lightweight gating module $h_G(\cdot)$ followed by a sigmoid to obtain a patch-wise gate $g_i = \sigma(h_G(r^T_i)) \in (0,1)$. This gate modulates the projected image residual in an element-wise manner:
\begin{equation}
    \bar{r}^{I}_i = g_i \odot \tilde{r}^{I}_i.
\end{equation}
The modulated residual feature is then passed through a lightweight scoring head $h_S(\cdot)$ to produce the text-guided score $S_{\mathrm{text}, i} = h_S(\bar{r}^{I}_i)$, which yields the text score map $S_{\mathrm{text}}=\{S_{\mathrm{text}, i}\}_{i=1}^{L}$.

In this way, the text branch does not act as an independent text-image similarity classifier. Instead, it computes text residual features from normal and abnormal textual states and uses them to guide patch-level image residuals. This design preserves the localization ability of image residual matching while incorporating semantic cues from defect-related textual descriptions.

\subsection{Anomaly Score Aggregation}
\label{sec:fusion}

The final patch-level anomaly score is obtained by merging the image score and the text-guided score:
\begin{equation}
    S_{\mathrm{final}}(i)
    =
    (1-\alpha)S_{\mathrm{image}}(i)
    +
    \alpha S_{\mathrm{text}}(i),
    \label{eq:fusion}
\end{equation}
where $\alpha=\sigma(a)$ is a learnable scalar weight that balances the two scores.
The patch-level score map is reshaped to the spatial resolution $h\times w$ and upsampled to the input image resolution $H\times W$ by bilinear interpolation to obtain the final localization map $S$. For image-level anomaly detection, we aggregate the highest-scoring patches by averaging the anomaly scores of the top $1\%$ patches in the map, which yields the image-level score $s$.

\subsection{Training Objective}
\label{sec:loss}

We train both pixel-level and image-level supervision. Given the predicted anom\-a\-ly map $S$ and the ground-truth mask $m^q$, we downsample $m^q$ to the patch resolution and supervise the localization output with a segmentation loss. Since anomalous regions typically occupy only a small fraction of the image, the pixel-level labels are highly imbalanced. We therefore combine a focal loss~\cite{lin2017focal}, which down-weights easy normal pixels and focuses training on hard anomalous ones, with a dice loss~\cite{li2020dice}, which directly optimizes the region-level overlap between the prediction and the ground-truth mask:
\begin{equation}
    \mathcal{L}_{\mathrm{seg}}
    =
    \mathcal{L}_{\mathrm{focal}}(S,m^q)
    +
    \mathcal{L}_{\mathrm{dice}}(S,m^q).
\end{equation}
The image-level anomaly score $s$ is supervised by binary cross-entropy:
\begin{equation}
    \mathcal{L}_{\mathrm{cls}}
    =
    \mathcal{L}_{\mathrm{bce}}(s,y^q).
\end{equation}

We further use a patch-text alignment loss to encourage patch features to be closer to the corresponding normal or abnormal text prototype. For each patch feature $f^q_i$, we compute a two-way text prediction over the mean projected normal and abnormal text features $\bar{t}^{n}$ and $\bar{t}^{a}$ (Sec.~\ref{sec:text_branch}):
\begin{equation}
    p_i = \operatorname{Softmax}\!\left(
    \left[\operatorname{cos}(f^q_i,\bar{t}^{n}),\;
    \operatorname{cos}(f^q_i,\bar{t}^{a})\right]/\tau'
    \right),
\end{equation}
where $\tau'$ is a temperature. The alignment loss is then defined as the mean two-way cross-entropy over all patches:
\begin{equation}
    \mathcal{L}_{\mathrm{align}}
    =
    \frac{1}{L}
    \sum_{i=1}^{L}
    \operatorname{CE}(p_i,\hat{m}^{q}_i),
\end{equation}
where $\hat{m}^{q}_i\in\{0,1\}$ is the downsampled patch-level anomaly label. This loss is part of the text branch rather than an auxiliary term, as it shapes the projected text space in which the patch-wise matching and the gate are computed.

The final objective is a weighted sum of the segmentation, classification, and alignment terms:
\begin{equation}
    \mathcal{L}
    =
    \mathcal{L}_{\mathrm{seg}}
    +
    \lambda_{\mathrm{cls}}\mathcal{L}_{\mathrm{cls}}
    +
    \lambda_{\mathrm{align}}\mathcal{L}_{\mathrm{align}}.
\end{equation}

\subsection{Inference}
\label{sec:inference}

At inference, all learned modules are frozen and no target-category adaptation is performed. For each unseen target category, we build the normal memory bank from the few reference images and instantiate the text prompts from the category name. The final localization map $S$ and image-level score $s$ are then obtained by the merging and aggregation procedure in Sec.~\ref{sec:fusion}.

\section{Experiments}
\label{sec:experiments}

\subsection{Experimental Setup}

\noindent\textbf{Datasets and Evaluation Protocol. }
We evaluate on MVTec AD~\cite{MVTec}, VisA~\cite{VisA}, and BraTS~\cite{brats} following the BMAD benchmark~\cite{bao2024bmad}. MVTec AD and VisA are industrial anomaly detection benchmarks, while BraTS is used to examine industrial-to-medical transfer. Following the generalist anomaly detection setting, we train the model on a source dataset and directly test it on a different target dataset without category-specific retraining. We consider three transfer settings: MVTec AD $\rightarrow$ VisA, VisA $\rightarrow$ MVTec AD, and MVTec AD $\rightarrow$ BraTS.

During training, normal images, abnormal images, and pixel-level ground-truth masks are used. During inference, only a few normal reference images from the target dataset are provided. We report results under 1-shot, 2-shot, and 4-shot settings. Image-level performance is measured by AUROC, AP, and F1-max, while pixel-level localization is evaluated by AUROC, F1-max, and PRO. All metrics are reported in percentage.

\noindent\textbf{Implementation Details. }
We use DINOv2-Txt~\cite{dinov2,dinotxt} with a ViT-L/14~\cite{vit, vit_reg} backbone as the frozen image-text foundation model. The image encoder and text encoder are fixed during training, and only lightweight modules for residual projection, scoring, and merging are optimized. All images are resized to $448 \times 448$. The model is trained for 20 epochs using AdamW~\cite{AdamW} with a learning rate of $1\times10^{-4}$, weight decay of $1\times10^{-4}$, and batch size of 32. The loss weights are set to $\lambda_{\mathrm{cls}}=1.0$ and $\lambda_{\mathrm{align}}=0.3$. All reported results are averaged over three random seeds.

\subsection{Main Results}

\begin{table*}[t]
\centering
\caption{Main results on MVTec AD and VisA under 1-, 2-, 4-shot settings. Results marked with $^*$ are based on our re-implementation under the same DINOv2-Txt backbone, while the remaining results are taken from~\cite{nagl}. Best results are in \textbf{bold}, and second-best results are \underline{underlined}.}
\label{tab:main_results}
\resizebox{\textwidth}{!}{
\begin{tabular}{c l ccc ccc ccc ccc}
\toprule
\multirow{3}{*}{Shot} & \multirow{3}{*}{Method}
& \multicolumn{6}{c}{MVTec AD}
& \multicolumn{6}{c}{VisA} \\
\cmidrule(lr){3-8} \cmidrule(lr){9-14}
& & \multicolumn{3}{c}{Image-level} & \multicolumn{3}{c}{Pixel-level}
& \multicolumn{3}{c}{Image-level} & \multicolumn{3}{c}{Pixel-level} \\
\cmidrule(lr){3-5} \cmidrule(lr){6-8}
\cmidrule(lr){9-11} \cmidrule(lr){12-14}
& & AUROC & AP & F1 & AUROC & F1 & PRO
& AUROC & AP & F1 & AUROC & F1 & PRO \\
\midrule

\multirow{6}{*}{1-shot}
& PatchCore & 83.4 & 92.2 & 90.5 & 92.0 & 50.4 & 79.7 & 79.9 & 82.8 & 81.7 & 95.4 & 38.0 & 80.5 \\
& WinCLIP & 93.1 & 96.5 & 93.7 & 95.2 & \underline{55.9} & 87.1 & 83.8 & 85.1 & 83.1 & 96.4 & \underline{41.3} & 85.1 \\
& AnomalyDINO$^*$ & 95.4 & 97.6 & 94.9 & 93.8 & 52.2 & 86.0 & 85.6 & 88.3 & 82.9 & 95.6 & 36.4 & 87.3 \\
& ResAD$^*$ & 92.8 & 95.5 & 93.9 & 94.9 & 55.4 & 92.1 & 89.3 & 91.2 & \underline{87.2} & 96.5 & 38.7 & 87.7 \\
& NAGL$^*$ & \underline{96.0} & \underline{97.8} & \underline{95.4} & \underline{95.8} & 55.7 & \textbf{92.7} & \underline{90.1} & \underline{91.6} & 86.6 & \underline{97.1} & 38.9 & \underline{89.7} \\

& \cellcolor{gray!15}\textbf{TIRAD(Ours)} & \cellcolor{gray!15}\textbf{97.1} & \cellcolor{gray!15}\textbf{98.4} & \cellcolor{gray!15}\textbf{96.2} & \cellcolor{gray!15}\textbf{95.9} & \cellcolor{gray!15}\textbf{56.2} & \cellcolor{gray!15}\underline{92.6} & \cellcolor{gray!15}\textbf{91.5} & \cellcolor{gray!15}\textbf{92.4} & \cellcolor{gray!15}\textbf{87.7} & \cellcolor{gray!15}\textbf{97.4} & \cellcolor{gray!15}\textbf{41.5} & \cellcolor{gray!15}\textbf{89.9} \\

\midrule
\multirow{7}{*}{2-shot}
& PatchCore & 86.3 & 93.8 & 92.0 & 93.3 & 53.0 & 82.3 & 81.6 & 84.8 & 82.5 & 96.1 & 41.0 & 82.6 \\
& WinCLIP & 94.4 & 97.0 & 94.4 & 96.0 & \textbf{58.4} & 88.4 & 84.6 & 85.8 & 83.0 & 96.8 & \textbf{43.5} & 86.2 \\
& InCTRL & 94.0 & 96.9 & -- & -- & -- & -- & 85.8 & 87.7 & -- & -- & -- & -- \\
& AnomalyDINO$^*$ & 95.9 & 97.8 & 95.4 & 94.2 & 54.8 & 92.5 & 87.4 & 89.8 & 84.4 & 96.8 & 34.2 & 88.1 \\
& ResAD$^*$ & 94.5 & 96.7 & 94.7 & 95.8 & 55.5 & 92.4 & 90.9 & 91.5 & \underline{87.8} & 96.7 & 39.5 & 89.9 \\
& NAGL$^*$ & \underline{96.7} & \underline{98.1} & \underline{96.1} & \underline{96.0} & \underline{56.9} & \textbf{93.1} & \underline{91.6} & \underline{92.6} & 87.7 & \underline{97.3} & 40.3 & \underline{90.7} \\

& \cellcolor{gray!15}\textbf{TIRAD(Ours)} & \cellcolor{gray!15}\textbf{97.6} & \cellcolor{gray!15}\textbf{98.7} & \cellcolor{gray!15}\textbf{97.1} & \cellcolor{gray!15}\textbf{96.2} & \cellcolor{gray!15} 55.9 & \cellcolor{gray!15}\underline{92.8} & \cellcolor{gray!15}\textbf{93.1} & \cellcolor{gray!15}\textbf{93.4} & \cellcolor{gray!15}\textbf{89.1} & \cellcolor{gray!15}\textbf{97.8} & \cellcolor{gray!15}\underline{42.1} & \cellcolor{gray!15}\textbf{90.8} \\

\midrule
\multirow{7}{*}{4-shot}
& PatchCore & 88.8 & 94.5 & 92.3 & 94.3 & 55.0 & 84.3 & 85.3 & 87.5 & 84.3 & 96.8 & \underline{43.9} & 84.9 \\
& WinCLIP & 95.2 & 97.3 & 94.7 & 96.2 & \textbf{59.5} & 89.0 & 87.3 & 88.8 & 84.2 & 97.2 & \textbf{47.0} & 87.6 \\
& InCTRL & 94.5 & 97.2 & -- & -- & -- & -- & 87.7 & 90.2 & -- & -- & -- & -- \\
& AnomalyDINO$^*$ & 96.9 & 98.3 & 96.2 & 94.3 & 53.4 & 86.8 & 89.1 & 91.3 & 85.9 & 97.0 & 35.2 & 88.6 \\
& ResAD$^*$ & 95.5 & 97.5 & 95.5 & 95.1 & 56.8 & 92.3 & 92.6 & 92.8 & 89.0 & 96.8 & 40.3 & 90.7 \\
& NAGL$^*$ & \underline{97.6} & \underline{98.5} & \underline{96.4} & \underline{96.2} & 57.6 & \underline{93.2} & \underline{93.1} & \underline{93.6} & \underline{89.1} & \underline{97.6} & 41.5 & \underline{91.2} \\

& \cellcolor{gray!15}\textbf{TIRAD(Ours)} & \cellcolor{gray!15}\textbf{98.2} & \cellcolor{gray!15}\textbf{99.2} & \cellcolor{gray!15}\textbf{97.7} & \cellcolor{gray!15}\textbf{96.5} & \cellcolor{gray!15}\underline{57.9} & \cellcolor{gray!15}\textbf{93.4} & \cellcolor{gray!15}\textbf{93.9} & \cellcolor{gray!15}\textbf{94.3} & \cellcolor{gray!15}\textbf{89.5} & \cellcolor{gray!15}\textbf{98.1} & \cellcolor{gray!15} 43.1 & \cellcolor{gray!15}\textbf{91.8} \\

\bottomrule
\end{tabular}
}
\end{table*}

\begin{table}[t]
\centering
\caption{
Cross-domain results on BraTS under 1-, 2-, and 4-shot settings.
All competing methods are reproduced under the same DINOv2-Txt backbone, where models are trained on MVTec AD and directly evaluated on BraTS without target-domain fine-tuning.
Best results are in \textbf{bold}, and second-best results are \underline{underlined}.
}
\label{tab:brain_results}
\resizebox{0.75\linewidth}{!}{
\begin{tabular}{c l ccc ccc}
\toprule
\multirow{2}{*}{Shot} & \multirow{2}{*}{Method}
& \multicolumn{3}{c}{Image-level}
& \multicolumn{3}{c}{Pixel-level} \\
\cmidrule(lr){3-5} \cmidrule(lr){6-8}
& & AUROC & AP & F1 & AUROC & F1 & PRO \\
\midrule

\multirow{4}{*}{1-shot}
& AnomalyDINO & 71.83 & 91.03 & 91.13 & 94.69 & 39.18 & \underline{76.43} \\
& ResAD & 71.95 & 90.47 & 91.07 & 95.32 & 37.47 & 76.29 \\
& NAGL & \underline{72.79} & \underline{91.05} & \underline{91.34} & \underline{95.36} & \underline{42.66} & \textbf{77.30} \\
& \cellcolor{gray!15}TIRAD(Ours) & \cellcolor{gray!15}\textbf{73.69} & \cellcolor{gray!15}\textbf{92.12} & \cellcolor{gray!15}\textbf{91.55} & \cellcolor{gray!15}\textbf{95.81} & \cellcolor{gray!15}\textbf{46.44} & \cellcolor{gray!15} 76.14 \\

\midrule
\multirow{4}{*}{2-shot}
& AnomalyDINO & 76.77 & 92.32 & 91.40 & 95.36 & 42.69 & 77.78 \\
& ResAD & 76.05 & 91.51 & 91.82 & \underline{96.19} & \underline{48.09} & 78.01 \\
& NAGL & \underline{76.96} & \underline{92.46} & \underline{91.89} & 95.62 & 44.51 & \underline{78.29} \\
& \cellcolor{gray!15}TIRAD(Ours) & \cellcolor{gray!15}\textbf{79.83} & \cellcolor{gray!15}\textbf{94.04} & \cellcolor{gray!15}\textbf{92.10} & \cellcolor{gray!15}\textbf{96.53} & \cellcolor{gray!15}\textbf{51.06} & \cellcolor{gray!15}\textbf{78.50} \\

\midrule
\multirow{4}{*}{4-shot}
& AnomalyDINO & 79.36 & 93.29 & 91.52 & 95.79 & 45.38 & 79.13 \\
& ResAD & 79.76 & 92.47 & 92.06 & 95.73 & 46.31 & 78.92 \\
& NAGL & \underline{80.53} & \underline{93.56} & \underline{92.23} & \underline{96.07} & \underline{47.23} & \underline{79.22} \\
& \cellcolor{gray!15}TIRAD(Ours) & \cellcolor{gray!15}\textbf{83.80} & \cellcolor{gray!15}\textbf{95.39} & \cellcolor{gray!15}\textbf{92.38} & \cellcolor{gray!15}\textbf{96.97} & \cellcolor{gray!15}\textbf{54.32} & \cellcolor{gray!15}\textbf{79.76} \\

\bottomrule
\end{tabular}
}
\end{table}

\noindent\textbf{Comparison with Existing Methods. }
Table~\ref{tab:main_results} reports the main results on MVTec AD and VisA. Against the strongest re-implemented baseline NAGL$^*$~\cite{nagl}, image-level AUROC improves by 1.1/0.9/0.6 points on MVTec AD and 1.4/1.5/\allowbreak 0.8 points on VisA under the 1-, 2-, and 4-shot settings,  with F1 gains of up to 1.4 points, while ResAD$^*$~\cite{resad} and AnomalyDINO$^*$~\cite{damm2024anomalydino} are outperformed by clear margins. These consistent gains indicate that text residual cues sharpen the object-level normal/abnormal decision, where purely visual residuals have largely saturated. For localization, our method attains the best pixel-level AUROC across all shots on both datasets, improves VisA pixel-level F1 over NAGL$^*$ by 2.6/1.8/1.6 points with the best PRO in every setting, and remains best or comparable on MVTec AD. The image-level improvement is thus the more decisive gap, while the clearest pixel-level gains emerge in the cross-domain setting discussed next.

\noindent\textbf{Cross-Domain Transfer. }
Table~\ref{tab:brain_results} evaluates industrial-to-medical transfer, where all methods share the same DINOv2-Txt backbone and are trained on MVTec AD without any BraTS fine-tuning.
Here the benefit of text residuals is most evident.
The image-level AUROC improves over the second-best method by 0.9, 2.9, and 3.3 points under the 1-, 2-, and 4-shot settings, and the pixel-level F1 by 3.8, 3.0, and 7.1 points.
Notably, the image-level margin grows steadily as more reference images become available, suggesting that semantic text residuals become increasingly effective at suppressing domain-specific visual variation once sufficient normal context is provided.

\subsection{Ablation Study}

\begin{table}[t]
\centering
\caption{
Ablation study on text residual modeling under the 1-shot setting on MVTec AD.
``matched'' directly uses the matched text feature, while ``residual'' uses the text residual feature computed with respect to the normal text anchor.
}
\label{tab:abl_text_residual}
\setlength{\tabcolsep}{6pt}
\begin{tabular}{lcccccc}
\toprule
& \multicolumn{3}{c}{Image-level} & \multicolumn{3}{c}{Pixel-level} \\
\cmidrule(lr){2-4} \cmidrule(lr){5-7}
\textbf{Configuration} & AUROC & AP & F1 & AUROC & F1 & PRO \\
\midrule
$S_{\text{image}}$                            & 93.79 & 96.60 & 94.77 & 95.61 & 54.21 & 91.48 \\
$S_{\text{image}}+S_{\text{text}}$ (matched)  & 95.47 & 97.50 & 95.56 & 95.67 & 54.82 & 91.98 \\
$S_{\text{image}}+S_{\text{text}}$ (residual) & \textbf{97.06} & \textbf{98.39} & \textbf{96.21} & \textbf{95.87} & \textbf{56.19} & \textbf{92.63} \\
\bottomrule
\end{tabular}
\end{table}

\begin{table*}[t]
\centering

\begin{minipage}[t]{0.49\textwidth}
\centering
\caption{Ablation study on the effect of the text and image projectors. We compare configurations that include or remove the learnable projectors, isolating the contribution of each projector and showing which configuration is most beneficial.}

\label{tab:ablation_adapter}
\resizebox{\linewidth}{!}{
\begin{tabular}{cc|ccc|ccc}
\toprule
\textbf{Text} & \textbf{Image}
& \multicolumn{3}{c|}{\textbf{Image-level}}
& \multicolumn{3}{c}{\textbf{Pixel-level}} \\
\cmidrule(lr){3-5} \cmidrule(lr){6-8}
\textbf{Proj} & \textbf{Proj}
& AUROC & AP & F1 & AUROC & F1 & PRO \\
\midrule
$\times$ & $\times$
& 95.18 & 97.27 & 95.08 & 95.76 & 55.39 & 92.19 \\
$\checkmark$ & $\times$
& 95.98 & 97.95 & 95.35 & 95.83 & 55.87 & 92.36 \\
$\times$ & $\checkmark$
& 96.75 & 98.22 & 96.10 & 95.85 & 56.18 & \textbf{92.66} \\
$\checkmark$ & $\checkmark$
& \textbf{97.06} & \textbf{98.39} & \textbf{96.21} & \textbf{95.87} & \textbf{56.19} & 92.63 \\
\bottomrule
\end{tabular}
}
\end{minipage}
\hfill
\begin{minipage}[t]{0.49\textwidth}
\centering
\caption{
Ablation study on the merge weight $\alpha$ in
$S = (1-\alpha)S_{\mathrm{image}} + \alpha S_{\mathrm{text}}$,
evaluated as a post-hoc sweep on the trained checkpoint
(learned $\alpha \approx 0.23$).
}
\label{tab:ablation_alpha}
\resizebox{\linewidth}{!}{
\begin{tabular}{c|ccc|ccc}
\toprule
& \multicolumn{3}{c|}{\textbf{Image-level}}
& \multicolumn{3}{c}{\textbf{Pixel-level}} \\
\cmidrule(lr){2-4} \cmidrule(lr){5-7}
$\alpha$ & AUROC & AP & F1 & AUROC & F1 & PRO \\
\midrule
0.0 & 93.79 & 96.60 & 94.77 & 95.61 & 54.21 & 91.48 \\
0.2 & \textbf{96.86} & \textbf{98.24} & \textbf{96.11} & {95.64} & \textbf{55.26} & \textbf{92.21} \\
0.4 & \underline{96.08} & \underline{97.81} & \underline{95.97} & \textbf{95.71} & \underline{54.60} & \underline{92.06} \\
0.6 & 96.00 & 97.61 & 95.77 & \underline{95.68} & 53.86 & 92.03 \\
0.8 & 95.33 & 96.95 & 95.16 & 95.43 & 52.79 & 91.72 \\
\bottomrule
\end{tabular}
}
\end{minipage}

\end{table*}
We conduct ablation studies to validate the contribution of the proposed components. Table~\ref{tab:abl_text_residual} analyzes the effect of text residual modeling. While the matched text feature already improves over the image-only baseline, the residual formulation yields a substantially larger gain, indicating that the semantic discrepancy relative to the normal anchor rather than the absolute matched embedding is the primary driver of the improvement.

Table~\ref{tab:ablation_adapter} studies the effect of the text projector and image projector. The results show that the text projector improves performance by mapping text residual features into a more discriminative space for anomaly scoring. The image projector also contributes by transforming image residual features before they are modulated by the text-derived gate. When both projectors are used together, the model achieves the best overall balance, indicating that the two projectors play complementary roles in aligning semantic and visual residual cues.

Table~\ref{tab:ablation_alpha} evaluates the effect of the merge weight $\alpha$. Relying only on the image residual score limits the benefit of semantic information, while assigning an excessively large weight to the text residual score weakens visual localization reliability. The best performance is obtained with a moderate merge weight, showing that text residual features are most effective when used as complementary guidance rather than as a replacement for image residual features.

\begin{figure}[t]
    \centering
    \includegraphics[width=1 \linewidth,trim={0 150pt 0 130pt}, clip]{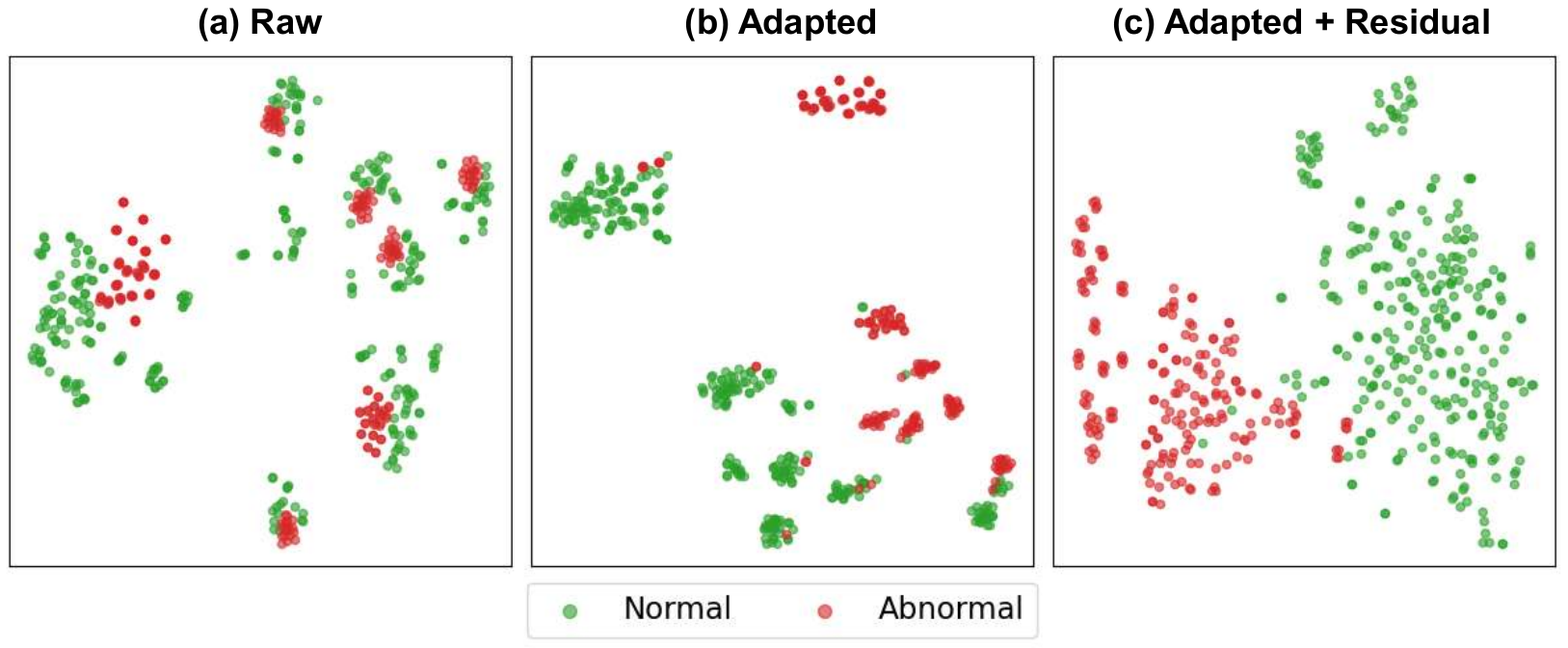}
    \caption{t-SNE visualization of text features across three stages: (a) raw features from the frozen encoder, (b) adapted features after the learned text projector, and (c) the resulting text residual features. Separation between normal and abnormal semantic states improves progressively from (a) to (c), and remains consistent across different object classes.}
    \label{fig:tsne_text_adapter}
\end{figure}

\begin{figure}[t]
    \centering
    \includegraphics[width=1.0\linewidth, trim={50 120pt 50 110pt}, clip]{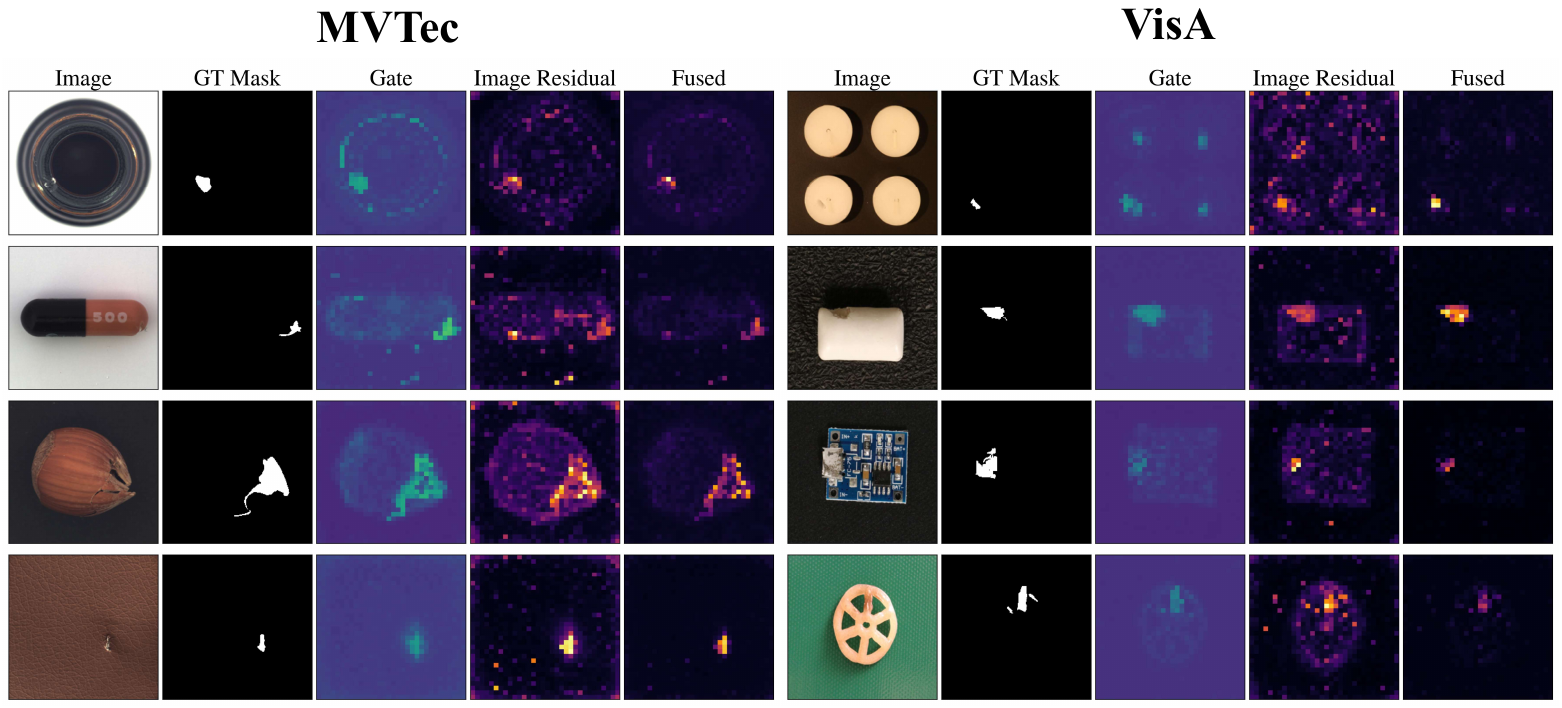}
    \caption{Qualitative anomaly localization on MVTec AD and VisA. From left to right: input image, ground-truth mask, text-residual gate, image residual response, and the fused anomaly map. The text-residual gate modulates the image residual, and the resulting fused map localizes anomalies more accurately than the image residual alone.}
    \label{fig:qual_mvtec_visa}
\end{figure}

\begin{figure}[t]
    \centering
    \includegraphics[width=1.0\linewidth, trim={50 70pt 50 100pt}, clip]{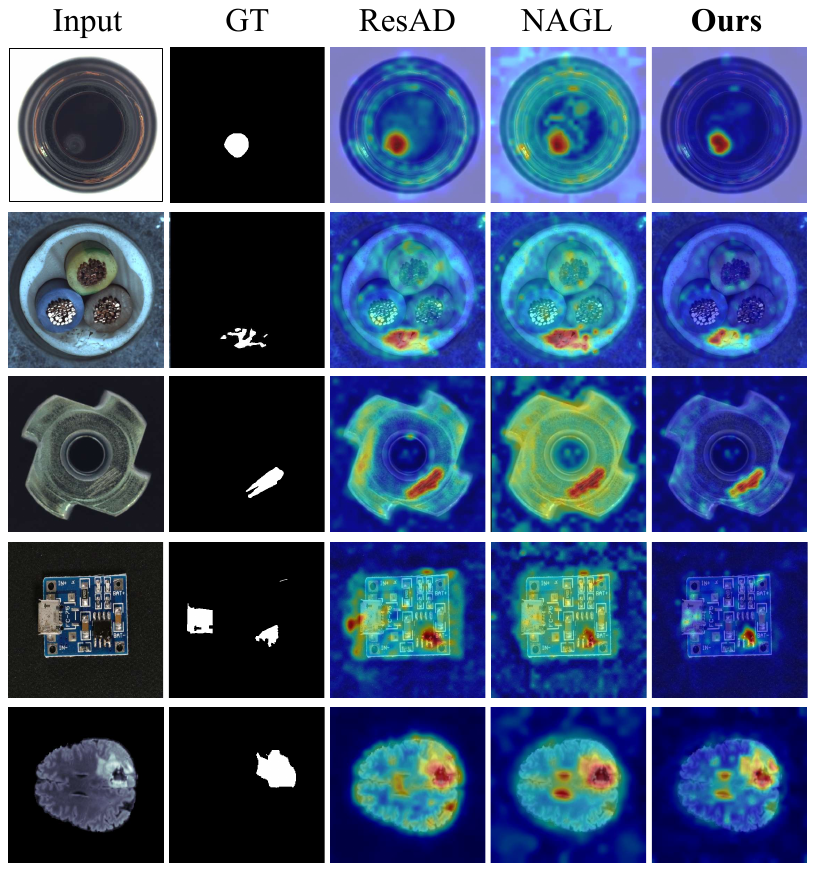}
    \caption{Qualitative comparison of anomaly maps produced by different methods, each following the setting in Sec.~\ref{sec:experiments}.}
    \label{fig:methods}
\end{figure}

\subsection{Analysis}
Fig.~\ref{fig:tsne_text_adapter} visualizes the text feature distributions across three stages: raw features, projector-adapted features, and text residual features. The raw text features from the frozen encoder show relatively weak separation between normal and abnormal semantic states, indicating that frozen textual embeddings are not directly optimized for anomaly discrimination. After the text projector, the distribution becomes more structured, suggesting that the projector reshapes textual representations toward the anomaly detection objective; notably, despite originating from different object classes, the adapted features form more coherent clusters organized by normal and abnormal states rather than by class. The text residual features exhibit the clearest separation of the three, demonstrating that residual semantic directions provide class-agnostic cues for anomaly discrimination.

\subsection{Qualitative Results}
Fig.~\ref{fig:qual_mvtec_visa} and Fig.~\ref{fig:methods} present qualitative anomaly localization results.
In Fig.~\ref{fig:qual_mvtec_visa}, the gate maps show that the text branch selectively modulates visually suspicious regions, suppressing irrelevant activations while emphasizing regions that are semantically consistent with abnormality.
By applying this gate to the image residual, the fused map yields more focused and coherent localization than the image residual alone, aligning more closely with the ground-truth anomalous regions and reducing false activations in normal background areas.

Fig.~\ref{fig:methods} further compares the anomaly maps of our method with those of competing methods.
Our results produce sharper and better-localized responses on the ground-truth anomalous regions, whereas the baselines tend to yield more diffuse activations or miss subtle defects.

\section{Conclusion}
\label{sec:conclusion}

We presented TIRAD, a generalist few-shot anomaly detection framework that complements conventional image residuals with text residuals. Rather than treating text as a static similarity anchor, our method computes residuals in the text feature space and uses them to gate patch-level image residuals, refining which visual discrepancies correspond to true anomalies while preserving the localization ability of visual matching. Built on a frozen backbone with only lightweight modules trained, it transfers to unseen categories without target-category fine-tuning, and yields consistent gains over strong same-backbone baselines across industrial benchmarks and industrial-to-medical transfer, with the largest improvements in the cross-domain regime. These results suggest that modeling semantic differences between normal and abnormal descriptions is a promising, underexplored direction for generalist anomaly detection.

A key limitation arises when a single patch contains both normal and abnormal regions: the mixed patch is assigned an ambiguous label that lies between the normal and abnormal textual states, which weakens the text residual and blurs boundary localization. Addressing such mixed patches, for example through finer-grained or soft patch-level supervision, is an important direction for future work.

\noindent\textbf{Acknowledgments.} 
This work was supported by the National Research Foundation (NRF) grant funded by the Korea government (MSIT) [RS-2025-00562400] and [RS-2022-NR068754].


%
%
\bibliographystyle{splncs04}
\bibliography{reference}

\begin{thebibliography}{10}
\providecommand{\url}[1]{\texttt{#1}}
\providecommand{\urlprefix}{URL }
\providecommand{\doi}[1]{https://doi.org/#1}

\bibitem{bao2024bmad}
Bao, J., Sun, H., Deng, H., He, Y., Zhang, Z., Li, X.: {Bmad}: Benchmarks for medical anomaly detection. In: Proceedings of the IEEE/CVF conference on computer vision and pattern recognition. pp. 4042--4053 (2024)

\bibitem{MVTec}
Bergmann, P., Fauser, M., Sattlegger, D., Steger, C.: {MVTec AD} -- a comprehensive real-world dataset for unsupervised anomaly detection. In: CVPR (2019)

\bibitem{adaclip}
Cao, Y., Zhang, J., Frittoli, L., Cheng, Y., Shen, W., Boracchi, G.: {AdaCLIP}: Adapting {CLIP} with hybrid learnable prompts for zero-shot anomaly detection. In: ECCV (2024)

\bibitem{chen2023VAND}
Chen, X., Han, Y., Zhang, J.: A zero-/few-shot anomaly classification and segmentation method for cvpr 2023 vand workshop challenge tracks 1\&2: 1st place on zero-shot ad and 4th place on few-shot ad. ArXiv  \textbf{abs/2305.17382} (2023)

\bibitem{spade}
Cohen, N., Hoshen, Y.: Sub-image anomaly detection with deep pyramid correspondences. ArXiv  \textbf{abs/2005.02357} (2020)

\bibitem{damm2024anomalydino}
Damm, S., Laszkiewicz, M., Lederer, J., Fischer, A.: Anomalydino: Boosting patch-based few-shot anomaly detection with dinov2. In: 2025 IEEE/CVF Winter Conference on Applications of Computer Vision (WACV). pp. 1319--1329. IEEE (2025)

\bibitem{vit_reg}
Darcet, T., Oquab, M., Mairal, J., Bojanowski, P.: Vision transformers need registers. ICLR  (2023)

\bibitem{padim}
Defard, T., Setkov, A., Loesch, A., Audigier, R.: {PaDiM}: A patch distribution modeling framework for anomaly detection and localization. In: ICPR (2021)

\bibitem{deng2022anomaly}
Deng, H., Li, X.: Anomaly detection via reverse distillation from one-class embedding. In: CVPR (2022)

\bibitem{vit}
Dosovitskiy, A.: An image is worth 16x16 words: Transformers for image recognition at scale. ICLR  (2020)

\bibitem{gao2025learning}
Gao, B.B.: Learning to detect multi-class anomalies with just one normal image prompt. In: ECCV (2025)

\bibitem{adaptclip}
Gao, B.B., Zhou, Y., Yan, J., Cai, Y., Zhang, W., Wang, M., Liu, J., Liu, Y., Wang, L., Wang, C.: Adaptclip: Adapting clip for universal visual anomaly detection. In: Proceedings of the AAAI Conference on Artificial Intelligence. vol.~40, pp. 4095--4103 (2026)

\bibitem{gu2024anomalygpt}
Gu, Z., Zhu, B., Zhu, G., Chen, Y., Tang, M., Wang, J.: Anomalygpt: Detecting industrial anomalies using large vision-language models. In: Proceedings of the AAAI conference on artificial intelligence. vol.~38, pp. 1932--1940 (2024)

\bibitem{gudovskiy2022cflow}
Gudovskiy, D., Ishizaka, S., Kozuka, K.: {CFLOW-AD}: Real-time unsupervised anomaly detection with localization via conditional normalizing flows. In: WACV (2022)

\bibitem{hou2026visualad}
Hou, Y., Li, P., Liu, Z., Wang, Y., Ruan, Y., Qiu, J., Xu, K.: Visualad: Language-free zero-shot anomaly detection via vision transformer. In: Proceedings of the IEEE/CVF Conference on Computer Vision and Pattern Recognition. pp. 21346--21356 (2026)

\bibitem{regad}
Huang, C., Guan, H., Jiang, A., Zhang, Y., Spratling, M., Wang, Y.F.: Registration based few-shot anomaly detection. In: European conference on computer vision. pp. 303--319. Springer (2022)

\bibitem{mvfa}
Huang, C., Jiang, A., Feng, J., Zhang, Y., Wang, X., Wang, Y.: Adapting visual-language models for generalizable anomaly detection in medical images. In: IEEE/CVF Conference on Computer Vision and Pattern Recognition (CVPR) (2024)

\bibitem{reconpatch}
Hyun, J., Kim, S., Jeon, G., Kim, S.H., Bae, K., Kang, B.: Reconpatch : Contrastive patch representation learning for industrial anomaly detection. In: IEEE Workshop/Winter Conference on Applications of Computer Vision (2023). \doi{10.1109/WACV57701.2024.00205}

\bibitem{winclip}
Jeong, J., Zou, Y., Kim, T., Zhang, D., Ravichandran, A., Dabeer, O.: {WinCLIP}: Zero-/few-shot anomaly classification and segmentation. In: CVPR (2023)

\bibitem{dinotxt}
Jose, C., Moutakanni, T., Kang, D., Baldassarre, F., Darcet, T., Xu, H., Li, D., Szafraniec, M., Ramamonjisoa, M., Oquab, M., et~al.: Dinov2 meets text: A unified framework for image-and pixel-level vision-language alignment. In: Proceedings of the Computer Vision and Pattern Recognition Conference. pp. 24905--24916 (2025)

\bibitem{promptad}
Li, X., Zhang, Z., Tan, X., Chen, C., Qu, Y., Xie, Y., Ma, L.: {PromptAD}: Learning prompts with only normal samples for few-shot anomaly detection. In: CVPR (2024)

\bibitem{li2020dice}
Li, X., Sun, X., Meng, Y., Liang, J., Wu, F., Li, J.: Dice loss for data-imbalanced nlp tasks. In: ACL (2020)

\bibitem{li2024musc}
Li, X., Huang, Z., Xue, F., Zhou, Y.: {MuSc}: Zero-shot industrial anomaly classification and segmentation with mutual scoring of the unlabeled images. In: ICLR (2024)

\bibitem{li2024onetonormal}
Li, Y., Zhang, S., Li, K., Lao, Q.: {One-to-Normal}: Anomaly personalization for few-shot anomaly detection. In: NeurIPS (2024)

\bibitem{lin2017focal}
Lin, T.Y., Goyal, P., Girshick, R., He, K., Doll{\'a}r, P.: Focal loss for dense object detection. In: ICCV (2017)

\bibitem{AdamW}
Loshchilov, I., Hutter, F.: Decoupled weight decay regularization. In: ICLR (2019)

\bibitem{IIPAD}
Lv, W., Su, Q., Xu, W.: One-for-all few-shot anomaly detection via instance-induced prompt learning. In: ICLR (2025)

\bibitem{brats}
Menze, B.H., Jakab, A., Bauer, S., Kalpathy-Cramer, J., Farahani, K., Kirby, J., et~al.: The multimodal brain tumor image segmentation benchmark {(BRATS)}. IEEE Trans. Med. Imaging  \textbf{34}(10),  1993--2024 (2014)

\bibitem{oh2025aulora}
Oh, S., Lee, S., Chae, S., Ro, Y.: Aulora: Anomaly understanding with low-rank adaptation for zero-shot anomaly detection. IEEE Access  (2025)

\bibitem{dinov2}
Oquab, M., Darcet, T., Moutakanni, T., Vo, H.V., Szafraniec, M., Khalidov, V., Fernandez, P., HAZIZA, D., Massa, F., El-Nouby, A., Assran, M., Ballas, N., Galuba, W., Howes, R., Huang, P.Y., Li, S.W., Misra, I., Rabbat, M., Sharma, V., Synnaeve, G., Xu, H., Jegou, H., Mairal, J., Labatut, P., Joulin, A., Bojanowski, P.: {DINO}v2: Learning robust visual features without supervision. Transactions on Machine Learning Research  (2024)

\bibitem{clip}
Radford, A., Kim, J.W., Hallacy, C., Ramesh, A., Goh, G., Agarwal, S., Sastry, G., Askell, A., Mishkin, P., Clark, J., Krueger, G., Sutskever, I.: Learning transferable visual models from natural language supervision. In: ICML (2021)

\bibitem{patchcore}
Roth, K., Pemula, L., Zepeda, J., Sch\"olkopf, B., Brox, T., Gehler, P.: Towards total recall in industrial anomaly detection. In: CVPR (2022)

\bibitem{sadikaj2025multiads}
Sadikaj, Y., Zhou, H., Halilaj, L., Schmid, S., Staab, S., Plant, C.: Multiads: Defect-aware supervision for multi-type anomaly detection and segmentation in zero-shot learning. In: Proceedings of the IEEE/CVF International Conference on Computer Vision. pp. 22978--22988 (2025)

\bibitem{medical_anomalydino}
Schulthess, N., Konukoglu, E.: Anomaly detection by clustering dino embeddings using a dirichlet process mixture. In: International Conference on Medical Image Computing and Computer-Assisted Intervention. pp. 46--56. Springer (2025)

\bibitem{madclip}
Shiri, M., Beyan, C., Murino, V.: Madclip: Few-shot medical anomaly detection with clip (2025), \url{https://arxiv.org/abs/2506.23810}

\bibitem{nagl}
Wang, Y., Wang, X., Gong, Y., Xiao, J.: Normal-abnormal guided generalist anomaly detection. NIPS  (2026)

\bibitem{uni_uninet}
Wei, S., Jiang, J., Xu, X.: Uninet: A contrastive learning-guided unified framework with feature selection for anomaly detection. In: Proceedings of the Computer Vision and Pattern Recognition Conference. pp. 9994--10003 (2025)

\bibitem{graphcore}
Xie, G., Wang, J., Liu, J., Jin, Y., Zheng, F.: Pushing the limits of fewshot anomaly detection in industry vision: {Graphcore}. In: ICLR (2023)

\bibitem{resad}
Yao, X., Chen, Z., Gao, C., Zhai, G., Zhang, C.: Resad: A simple framework for class generalizable anomaly detection. Advances in Neural Information Processing Systems  \textbf{37},  125287--125311 (2024)

\bibitem{yao2026adpretrain}
Yao, X., Luo, Y., Qian, Z., Zhang, C.: Adpretrain: Advancing industrial anomaly detection via anomaly representation pretraining. Advances in Neural Information Processing Systems  \textbf{38},  122977--123005 (2026)

\bibitem{you2022unified}
You, Z., Cui, L., Shen, Y., Yang, K., Lu, X., Zheng, Y., Le, X.: A unified model for multi-class anomaly detection. In: NeurIPS (2022)

\bibitem{yu2021fastflow}
Yu, J., Zheng, Y., Wang, X., Li, W., Wu, Y., Zhao, R., Wu, L.: {FastFlow}: Unsupervised anomaly detection and localization via 2d normalizing flows. arXiv preprint arXiv:2111.07677  (2021)

\bibitem{zavrtanik2021draem}
Zavrtanik, V., Kristan, M., Sko{\v{c}}aj, D.: {DRAEM}-a discriminatively trained reconstruction embedding for surface anomaly detection. In: ICCV (2021)

\bibitem{zhang2023prototypical}
Zhang, H., Wu, Z., Wang, Z., Chen, Z., Jiang, Y.G.: Prototypical residual networks for anomaly detection and localization. In: CVPR (2023)

\bibitem{mediclip}
Zhang, X., Xu, M., Qiu, D., Yan, R., Lang, N., Zhou, X.: Mediclip: Adapting clip for few-shot medical image anomaly detection. In: International Conference on Medical Image Computing and Computer-Assisted Intervention. pp. 458--468. Springer (2024)

\bibitem{Zhou_2022_CVPR_CoCoOp}
Zhou, K., Yang, J., Loy, C.C., Liu, Z.: Conditional prompt learning for vision-language models. In: CVPR (2022)

\bibitem{zhou2022CoOp}
Zhou, K., Yang, J., Loy, C.C., Liu, Z.: Learning to prompt for vision-language models. International Journal of Computer Vision  \textbf{130}(9),  2337--2348 (2022)

\bibitem{anomalyclip}
Zhou, Q., Pang, G., Tian, Y., He, S., Chen, J.: Anomaly{CLIP}: Object-agnostic prompt learning for zero-shot anomaly detection. In: ICLR (2024)

\bibitem{inctrl}
Zhu, J., Pang, G.: Toward generalist anomaly detection via in-context residual learning with few-shot sample prompts. In: CVPR (2024)

\bibitem{VisA}
Zou, Y., Jeong, J., Pemula, L., Zhang, D., Dabeer, O.: Spot-the-difference self-supervised pre-training for anomaly detection and segmentation. In: ECCV (2022)

\end{thebibliography}

\clearpage
\appendix
\setlength{\LTcapwidth}{\textwidth}
\providecommand{\std}[1]{{\scriptsize$\pm$#1}}

\renewcommand{\thesection}{\Alph{section}}
\renewcommand{\thetable}{\Alph{section}\arabic{table}}
\renewcommand{\thefigure}{\Alph{section}\arabic{figure}}

\setcounter{section}{0}
\setcounter{table}{0}
\setcounter{figure}{0}

\begin{center}
{\Large\bfseries Supplementary Material\par}
\vspace{0.5em}
{\normalsize TIRAD: Text-guided Image Residual Learning for Generalist Anomaly Detection\par}
\end{center}
\vspace{1em}

\section{Prompt Taxonomy}
\label{supp:prompt}

We provide the full set of text prompts used by the text branch.
For each domain, we use a list of normal-state prompts and,
for every defect type, a list of abnormal-state prompts. All prompts are wrapped by
the shared template ensemble below, and the resulting embeddings are averaged per
state before projection.

\subsection{Provenance}
\label{supp:prompt-provenance}

Prompt engineering is not a contribution of this work. For the industrial domains
(MVTec AD and VisA), we directly adopt the defect-aware prompt taxonomy of
MultiADS~\cite{sadikaj2025multiads}: the normal-state list, the template ensemble, and the
per-defect-type prompt sets are taken from its knowledge base for anomalies without
modification. This choice is deliberate: reusing an off-the-shelf prompt set isolates
our contribution to the residual formulation and the text-guided gate rather than to
prompt design.

For the medical domain (BraTS), no comparable public prompt set covers brain MRI, so
we construct the prompts ourselves, mirroring the structure of the industrial taxonomy
(one normal-state list plus per-anomaly-type descriptor lists). These descriptors
denote generic semantic states that may appear in brain MRI; they are \emph{not}
diagnostic labels and are not derived from any prior medical anomaly-detection method.
In the BraTS-based benchmark the ground-truth anomaly is tumor presence, whereas the
abnormal prompts intentionally span a superset of brain-anomaly descriptors so that the
patch-level soft matching can select the closest semantic state.

\subsection{Template Ensemble}
\label{supp:prompt-templates}

Every state prompt is expanded by the following 35-way template ensemble
(shared across all domains, following~\cite{sadikaj2025multiads}); \texttt{\{\}} is replaced by the
filled state string:

\begin{quote}\footnotesize\raggedright
\texttt{a bad photo of a [cls].}, \texttt{a low resolution photo of the [cls].}, \texttt{a bad photo of the [cls].}, \texttt{a cropped photo of the [cls].}, \texttt{a bright photo of a [cls].}, \texttt{a dark photo of the [cls].}, \texttt{a photo of my [cls].}, \texttt{a photo of the cool [cls].}, \texttt{a close-up photo of a [cls].}, \texttt{a black and white photo of the [cls].}, \texttt{a bright photo of the [cls].}, \texttt{a cropped photo of a [cls].}, \texttt{a jpeg corrupted photo of a [cls].}, \texttt{a blurry photo of the [cls].}, \texttt{a photo of the [cls].}, \texttt{a good photo of the [cls].}, \texttt{a photo of one [cls].}, \texttt{a close-up photo of the [cls].}, \texttt{a photo of a [cls].}, \texttt{a low resolution photo of a [cls].}, \texttt{a photo of a large [cls].}, \texttt{a blurry photo of a [cls].}, \texttt{a jpeg corrupted photo of the [cls].}, \texttt{a good photo of a [cls].}, \texttt{a photo of the small [cls].}, \texttt{a photo of the large [cls].}, \texttt{a black and white photo of a [cls].}, \texttt{a dark photo of a [cls].}, \texttt{a photo of a cool [cls].}, \texttt{a photo of a small [cls].}, \texttt{there is a [cls] in the scene.}, \texttt{there is the [cls] in the scene.}, \texttt{this is a [cls] in the scene.}, \texttt{this is the [cls] in the scene.}, \texttt{this is one [cls] in the scene.}
\end{quote}

\subsection{Normal-State Prompts}
\label{supp:prompt-normal}

\begin{table}[t]
\centering
\caption{Normal-state prompt templates for the three evaluation domains. The industrial lists follow the taxonomy of MultiADS~\cite{sadikaj2025multiads}; the medical list is constructed by us (Sec.~\ref{supp:prompt-provenance}). \texttt{[cls]} is replaced by the category name.}
\label{supp:tab:normal}
\small
\setlength{\tabcolsep}{4pt}
\begin{tabular}{@{}p{0.05\linewidth} p{0.30\linewidth} p{0.30\linewidth} p{0.30\linewidth}@{}}
\toprule
& \textbf{MVTec AD} & \textbf{VisA} & \textbf{BraTS (medical)} \\
\midrule
1 & [cls] & [cls] & [cls] \\
2 & flawless [cls] & flawless [cls] & healthy [cls] \\
3 & perfect [cls] & perfect [cls] & normal [cls] \\
4 & unblemished [cls] & unblemished [cls] & intact [cls] \\
5 & [cls] without flaw & [cls] without flaw & [cls] without abnormality \\
6 & [cls] without defect & [cls] without defect & [cls] without lesion \\
7 & [cls] without damage & [cls] without damage & [cls] without any anomaly \\
8 & [cls] in ideal condition & [cls] in pristine condition & [cls] in normal condition \\
9 & [cls] with immaculate quality & [cls] with no imperfections & [cls] with normal appearance \\
10 & [cls] without any imperfections & [cls] with ideal quality & [cls] with no pathological region \\
\bottomrule
\end{tabular}
\end{table}

\subsection{Industrial Defect-Aware Prompts}
\label{supp:prompt-industrial}

Tables~\ref{supp:tab:mvtec-defect} and~\ref{supp:tab:visa-defect} list the complete
defect-aware prompts for MVTec AD and VisA. Each dataset uses a superset of defect
types across all of its categories; at test time the category name fills \texttt{[cls]}.

\begin{longtable}{@{}p{0.18\linewidth} p{0.76\linewidth}@{}}
\caption{\footnotesize Defect-aware prompts for MVTec AD (20 defect states, superset over 15 categories), adopted from the MultiADS knowledge base~\cite{sadikaj2025multiads}.}\label{supp:tab:mvtec-defect}\\
\toprule
\textbf{Defect type} & \textbf{Prompts (\texttt{[cls]} = category name)} \\
\midrule
\endfirsthead
\toprule \textbf{Defect type} & \textbf{Prompts} \\ \midrule \endhead
\bottomrule \endfoot
\texttt{bent} & [cls] has a bent defect; flawed [cls] with a bent; a bend found in [cls]; [cls] with noticeable bending; a bent edge on [cls]; [cls] has a slight curve defect \\
\addlinespace
\texttt{broken} & [cls] has a broken defect; flawed [cls] with breakage; visible breakage on [cls]; [cls] with broken areas; [cls] shows signs of breaking; cracked or broken spots on [cls] \\
\addlinespace
\texttt{color} & [cls] has a color defect; inconsistent color on [cls]; [cls] with color discrepancies; [cls] has a noticeable color difference; [cls] with irregular coloring; [cls] has off-color patches \\
\addlinespace
\texttt{combined} & [cls] has a combined defect; multiple issues with [cls]; [cls] with mixed defects; [cls] showing multiple imperfections; several flaws found on [cls]; [cls] with combined defect types \\
\addlinespace
\texttt{contamination} & [cls] has a contamination defect; foreign particles on [cls]; [cls] is contaminated; [cls] contains contaminants; [cls] has impurity issues; traces of contamination on [cls] \\
\addlinespace
\texttt{crack} & [cls] has a crack defect; a crack is present on [cls]; cracked area on [cls]; [cls] with noticeable cracking; fine cracks found on [cls]; [cls] shows surface cracks \\
\addlinespace
\texttt{cut} & [cls] has a cut defect; cut marks on [cls]; [cls] with visible cuts; a cut detected on [cls]; [cls] is sliced or cut; surface cut seen on [cls] \\
\addlinespace
\texttt{fabric} & [cls] has a fabric defect; [cls] has a fabric border defect; [cls] has a fabric interior defect; fabric quality issues on [cls]; [cls] with textile irregularities; fabric borders on [cls] show defects \\
\addlinespace
\texttt{faulty imprint} & [cls] has a faulty imprint defect; [cls] has a print defect; incorrect printing on [cls]; misaligned print on [cls]; printing errors present on [cls]; [cls] has a blurred print defect \\
\addlinespace
\texttt{glue} & [cls] has a glue defect; [cls] has a glue strip defect; excess glue on [cls]; [cls] with uneven glue application; [cls] has visible glue spots; misplaced glue seen on [cls] \\
\addlinespace
\texttt{hole} & [cls] has a hole defect; a hole on [cls]; visible hole on [cls]; [cls] with punctures; small hole found in [cls]; perforations present on [cls] \\
\addlinespace
\texttt{missing} & [cls] has a missing defect; flawed [cls] with something missing; [cls] has missing components; missing parts on [cls]; [cls] shows absent pieces; certain parts missing from [cls] \\
\addlinespace
\texttt{poke} & [cls] has a poke defect; [cls] has a poke insulation defect; visible poke mark on [cls]; [cls] has puncture marks; a poke flaw on [cls]; small poke defect on [cls] \\
\addlinespace
\texttt{rough} & [cls] has a rough defect; rough texture on [cls]; uneven surface on [cls]; [cls] is coarser than expected; surface roughness seen on [cls]; texture defects on [cls] \\
\addlinespace
\texttt{scratch} & [cls] has a scratch defect; flawed [cls] with a scratch; visible scratches on [cls]; [cls] with surface scratches; minor scratches seen on [cls]; [cls] shows scratch marks \\
\addlinespace
\texttt{squeeze} & [cls] has a squeeze defect; flawed [cls] with a squeeze; squeezed area on [cls]; [cls] has compression marks; [cls] appears squeezed; flattened areas on [cls] \\
\addlinespace
\texttt{thread} & [cls] has a thread defect; flawed [cls] with a thread; loose threads on [cls]; [cls] has visible threads; untrimmed threads on [cls]; threads sticking out on [cls] \\
\addlinespace
\texttt{liquid} & [cls] has a liquid defect; flawed [cls] with liquid; [cls] with oil; liquid marks on [cls]; [cls] with liquid residue; stains from liquid on [cls] \\
\addlinespace
\texttt{misplaced} & [cls] has a misplaced defect; flawed [cls] with misplacing; [cls] shows misalignment; misplaced parts on [cls]; [cls] with incorrect positioning; positioning defects on [cls] \\
\addlinespace
\texttt{damaged} & [cls] has a damaged defect; flawed [cls] with damage; [cls] with visible damage; damaged areas on [cls]; physical damage seen on [cls]; noticeable wear on [cls] \\
\addlinespace
\end{longtable}

\begin{longtable}{@{}p{0.18\linewidth} p{0.76\linewidth}@{}}
\caption{\footnotesize Defect-aware prompts for VisA (18 defect states, superset over 12 categories), adopted from the MultiADS knowledge base~\cite{sadikaj2025multiads}.}\label{supp:tab:visa-defect}\\
\toprule
\textbf{Defect type} & \textbf{Prompts (\texttt{[cls]} = category name)} \\
\midrule
\endfirsthead
\toprule \textbf{Defect type} & \textbf{Prompts} \\ \midrule \endhead
\bottomrule \endfoot
\texttt{damage} & [cls] has a damaged defect; flawed [cls] with damage; [cls] shows signs of damage; damage found on [cls]; [cls] with visible wear and tear; [cls] with structural damage \\
\addlinespace
\texttt{scratch} & [cls] has a scratch defect; flawed [cls] with a scratch; scratches visible on [cls]; [cls] has surface scratches; small scratches found on [cls]; [cls] with scratch marks \\
\addlinespace
\texttt{breakage} & [cls] with a breakage defect; broken [cls]; [cls] with broken defect; [cls] shows breakage; broken or cracked areas on [cls]; visible breakage on [cls] \\
\addlinespace
\texttt{burnt} & [cls] with a burnt defect; [cls] shows burn marks; burnt areas on [cls]; [cls] with signs of burning; scorch marks on [cls]; [cls] appears slightly burnt \\
\addlinespace
\texttt{weird wick} & [cls] with a weird wick defect; [cls] has an unusual wick; the wick on [cls] appears odd; [cls] with a strangely shaped wick; irregular wick found on [cls]; odd wick defect on [cls] \\
\addlinespace
\texttt{stuck} & [cls] with a stuck defect; [cls] stuck together; [cls] has stuck parts; adhesive issue causing [cls] to stick; [cls] is partially stuck; [cls] with adhesion defect \\
\addlinespace
\texttt{crack} & [cls] with a crack defect; [cls] has a visible crack; cracked areas on [cls]; [cls] with surface cracking; fine cracks found on [cls]; [cls] shows crack lines \\
\addlinespace
\texttt{wrong place} & [cls] with defect that something on wrong place; [cls] has a misplaced defect; flawed [cls] with misplacing; misaligned part on [cls]; [cls] shows parts out of place; misplacement detected on [cls] \\
\addlinespace
\texttt{partical} & [cls] with particles defect; [cls] has foreign particles; small particles on [cls]; [cls] with unwanted particles; contaminants found on [cls]; [cls] with visible particles \\
\addlinespace
\texttt{bubble} & [cls] with bubbles defect; bubbles seen on [cls]; [cls] with bubble marks; air bubbles in [cls]; [cls] contains bubble defects; small bubbles on [cls] surface \\
\addlinespace
\texttt{melded} & [cls] with melded defect; melded parts on [cls]; [cls] has fused areas; fused spots on [cls]; melded areas on [cls]; [cls] with melded material \\
\addlinespace
\texttt{hole} & [cls] has a hole defect; a hole on [cls]; visible hole on [cls]; [cls] has small punctures; [cls] shows perforations; hole present on [cls] \\
\addlinespace
\texttt{melt} & [cls] with melt defect; melted areas on [cls]; [cls] shows melting; signs of melting on [cls]; [cls] with melted spots; [cls] has a melted appearance \\
\addlinespace
\texttt{bent} & [cls] has a bent defect; flawed [cls] with a bent; bent areas on [cls]; [cls] with visible bending; [cls] shows curvature issues; noticeable bend in [cls] \\
\addlinespace
\texttt{spot} & [cls] with spot defect; spots visible on [cls]; flawed [cls] with spots; [cls] with visible spotting; [cls] shows small spots; surface spots on [cls] \\
\addlinespace
\texttt{extra} & [cls] with extra thing; [cls] has a defect with extra thing; extra material on [cls]; [cls] contains additional pieces; [cls] with extra component defect; unwanted additions on [cls] \\
\addlinespace
\texttt{chip} & [cls] with chip defect; [cls] with fragment broken defect; chipped areas on [cls]; [cls] with chipped parts; broken fragments on [cls]; chip marks found on [cls] \\
\addlinespace
\texttt{missing} & [cls] with a missing defect; flawed [cls] with something missing; [cls] has missing parts; missing components on [cls]; absent pieces in [cls]; [cls] is incomplete \\
\addlinespace
\end{longtable}

\subsection{Medical Anomaly-State Prompts}
\label{supp:prompt-medical}

Table~\ref{supp:tab:brain-defect} lists the brain-MRI prompts described in
Sec.~\ref{supp:prompt-provenance}. Here \texttt{[cls]} is fixed to \texttt{brain}.

\begin{longtable}{@{}p{0.20\linewidth} p{0.74\linewidth}@{}}
\caption{Anomaly-state prompts for the BraTS (brain MRI) domain. These are semantic state descriptors constructed by us, mirroring the industrial taxonomy; they are not diagnostic labels and are not adopted from prior medical work.}\label{supp:tab:brain-defect}\\
\toprule
\textbf{Defect type} & \textbf{Prompts (\texttt{[cls]} = category name)} \\
\midrule
\endfirsthead
\toprule \textbf{Defect type} & \textbf{Prompts} \\ \midrule \endhead
\bottomrule \endfoot
\texttt{lesion} & [cls] has a lesion; a lesion is present on [cls]; abnormal lesion in [cls]; [cls] with a visible lesion; lesion region found in [cls]; [cls] shows a focal lesion \\
\addlinespace
\texttt{tumor} & [cls] has a tumor; a tumor is present in [cls]; tumor region in [cls]; [cls] with an abnormal mass; [cls] shows tumor tissue; abnormal growth detected in [cls] \\
\addlinespace
\texttt{hemorrhage} & [cls] has a hemorrhage; bleeding region in [cls]; [cls] with hemorrhagic area; [cls] shows signs of bleeding; hemorrhage found in [cls]; abnormal bleeding in [cls] \\
\addlinespace
\texttt{edema} & [cls] has edema; edema region in [cls]; [cls] with swelling; [cls] shows abnormal fluid accumulation; swollen area in [cls]; edematous region found in [cls] \\
\addlinespace
\texttt{mass} & [cls] has an abnormal mass; a mass is present in [cls]; mass region in [cls]; [cls] with an abnormal mass effect; [cls] shows a mass; abnormal mass detected in [cls] \\
\addlinespace
\texttt{anomaly} & [cls] has an anomaly; an abnormal region in [cls]; [cls] with abnormal tissue; [cls] shows an abnormality; abnormal area found in [cls]; [cls] with a pathological region \\
\addlinespace
\end{longtable}

\section{Implementation Details}
\label{supp:impl}

\subsection{Backbone and training}
We use DINOv2-Txt~\cite{dinotxt} with a ViT-L/14 backbone as the frozen image--text
foundation model; both the image encoder $\phi(\cdot)$ and the text encoder are fixed,
and only the lightweight residual-projection, gating, scoring, and merge modules are
trained. All images are resized to $448\times448$. We train for $20$ epochs with
AdamW~\cite{AdamW} (learning rate $1\times10^{-4}$, weight decay $1\times10^{-4}$,
batch size $32$). The loss weights are set to $\lambda_{\mathrm{cls}}=1.0$ and
$\lambda_{\mathrm{align}}=0.3$. All reported numbers are averaged over three random seeds.

\subsection{Text prompts}
For each category we encode the $P$ normal-state prompts (Table~\ref{supp:tab:normal})
and average them into a single normal text anchor $t^{n}_{\mathrm{anc}}$ used for the
text-residual subtraction. In parallel, each defect type is encoded and averaged into
one abnormal candidate, giving $M$ abnormal candidates ($M=20$ for MVTec AD, $M=18$ for
VisA, $M=6$ for BraTS). The patch-wise soft matching in Sec.~3.3 is performed over the
union of the projected normal candidates and the $M$ projected abnormal candidates,
with softmax temperature $\tau=0.07$. 

\subsection{Module specification}
The image and text projectors $g_I,g_T$ are lightweight linear maps in the backbone
embedding space, with $g_T$ initialized to the identity. The gate $h_G$
is a patch-wise linear head ($C\!\to\!1$) followed by a sigmoid, with bias initialized
to $-2.0$; the scoring head $h_S$ maps the gated residual to a scalar.

\subsection{Score aggregation and objective}
The two score maps are merged as
$S_{\mathrm{final}}=(1-\alpha)S_{\mathrm{image}}+\alpha S_{\mathrm{text}}$ with a
learnable scalar $\alpha=\sigma(a)$ initialized to $0.2$; after training it converges to
$\alpha\approx0.23$. The merge-weight ablation in the main paper is a post-hoc sweep of
$\alpha$ on this single trained checkpoint rather than a re-training sweep, which isolates
it from optimization variance; the learned value matches the optimum measured there.
Image-level scores use the mean of the top $1\%$ patch scores. The training objective follows Sec.~3.5,
$\mathcal{L}=\mathcal{L}_{\mathrm{seg}}+\lambda_{\mathrm{cls}}\mathcal{L}_{\mathrm{cls}}
+\lambda_{\mathrm{align}}\mathcal{L}_{\mathrm{align}}$, with
$\mathcal{L}_{\mathrm{seg}}=\mathcal{L}_{\mathrm{focal}}+\mathcal{L}_{\mathrm{dice}}$.

\section{Dataset Details}
\label{supp:data}

We evaluate on two industrial benchmarks and one medical benchmark under the generalist
setting, training on a source dataset and testing on a different target dataset without
category-specific retraining. \textbf{MVTec AD}~\cite{MVTec} contains $15$ object and
texture categories. \textbf{VisA}~\cite{VisA} contains $12$ object categories.
\textbf{BraTS}~\cite{brats} is a publicly available brain-MRI benchmark; we use its
reorganization into a binary anomaly-detection task (normal slices vs.\ slices
containing tumor)~\cite{bao2024bmad}. We follow the patient-level split of BraTS2021
provided by BMAD, so no patient appears in both the training and test sets.
All three datasets are publicly available and contain no personally identifying
information.
We consider three transfer settings: MVTec AD~$\rightarrow$~VisA,
VisA~$\rightarrow$~MVTec AD, and MVTec AD~$\rightarrow$~BraTS. During training, normal
images, abnormal images, and pixel-level masks are available; at inference only a few
normal reference images from the target dataset are provided, under $1$-, $2$-, and
$4$-shot settings. Image-level performance uses AUROC, AP, and F1-max; pixel-level
localization uses AUROC, F1-max, and PRO.

\section{Per-Class Results}
\label{supp:perclass}

For completeness we report per-category results (mean$\pm$std) under the $1$-, $2$-,
and $4$-shot settings. MVTec AD and VisA are given in Tables~\ref{tab:supp_mvtec_percat}
and~\ref{tab:supp_visa_percat}, respectively; each table stacks the 1-, 2-, and 4-shot
settings.
BraTS is single-class under the BMAD reorganization~\cite{bao2024bmad} and thus has no
per-category breakdown; its dataset-level result is given in
Table~\ref{tab:supp_brats_percat}. Unless noted otherwise, statistics are computed as
the sample standard deviation over three seeds; for the \textbf{Mean} rows, the
per-category scores are first averaged within each seed and the standard deviation is
then taken across the three seed-level averages.

{\scriptsize
\captionsetup{font=footnotesize}
\setlength{\tabcolsep}{2pt}
\begin{longtable}{@{}c l ccc ccc@{}}
\caption{Per-category results on MVTec AD under the 1-, 2-, and 4-shot settings, averaged over three random seeds (mean$\pm$std).}\label{tab:supp_mvtec_percat}\\
\toprule
& & \multicolumn{3}{c}{Image-level} & \multicolumn{3}{c}{Pixel-level}\\
\cmidrule(lr){3-5}\cmidrule(lr){6-8}
& Category & AUROC & AP & F1 & AUROC & F1 & PRO\\
\midrule
\endfirsthead
\multicolumn{8}{c}{\tablename~\thetable\ -- continued from previous page}\\
\toprule
& & \multicolumn{3}{c}{Image-level} & \multicolumn{3}{c}{Pixel-level}\\
\cmidrule(lr){3-5}\cmidrule(lr){6-8}
& Category & AUROC & AP & F1 & AUROC & F1 & PRO\\
\midrule
\endhead
\bottomrule
\endfoot
\multirow{16}{*}{1-shot} & bottle & 99.92\std{0.11} & 99.97\std{0.06} & 99.60\std{0.57} & 98.78\std{0.31} & 79.62\std{0.48} & 96.64\std{0.08}\\ & cable & 95.57\std{0.44} & 97.90\std{0.41} & 91.01\std{0.14} & 93.55\std{0.22} & 65.37\std{0.21} & 86.63\std{0.91}\\ & capsule & 92.96\std{0.37} & 98.97\std{0.35} & 94.93\std{0.33} & 95.37\std{0.06} & 45.41\std{0.56} & 95.33\std{0.23}\\ & carpet & 99.80\std{0.25} & 99.91\std{0.07} & 99.15\std{0.41} & 99.03\std{0.02} & 61.46\std{0.52} & 97.56\std{0.21}\\ & grid & 100.00\std{0.00} & 100.00\std{0.00} & 100.00\std{0.00} & 99.18\std{0.21} & 43.34\std{0.37} & 96.07\std{0.14}\\ & hazelnut & 99.98\std{0.03} & 99.99\std{0.02} & 99.65\std{0.50} & 99.25\std{0.04} & 74.17\std{1.48} & 96.72\std{0.80}\\ & leather & 100.00\std{0.00} & 100.00\std{0.00} & 100.00\std{0.00} & 99.03\std{0.04} & 43.00\std{0.04} & 98.70\std{0.01}\\ & metal\_nut & 99.71\std{0.07} & 99.93\std{0.13} & 98.91\std{0.01} & 95.23\std{0.98} & 70.59\std{4.78} & 93.35\std{0.91}\\ & pill & 98.70\std{0.18} & 99.77\std{0.14} & 97.16\std{0.45} & 91.48\std{0.16} & 50.22\std{0.46} & 96.47\std{0.16}\\ & screw & 83.85\std{0.23} & 93.29\std{0.57} & 89.02\std{0.42} & 98.25\std{0.21} & 49.90\std{0.11} & 93.56\std{1.07}\\ & tile & 100.00\std{0.00} & 100.00\std{0.00} & 100.00\std{0.00} & 95.64\std{0.25} & 64.88\std{0.12} & 90.48\std{0.35}\\ & toothbrush & 94.61\std{0.17} & 97.93\std{0.47} & 93.22\std{0.16} & 96.92\std{0.81} & 35.30\std{4.14} & 93.35\std{1.07}\\ & transistor & 92.15\std{0.38} & 88.76\std{0.94} & 84.14\std{0.49} & 87.34\std{1.97} & 43.72\std{2.98} & 68.98\std{2.38}\\ & wood & 99.30\std{0.62} & 99.78\std{0.19} & 97.87\std{1.70} & 93.76\std{0.73} & 62.61\std{0.59} & 94.28\std{0.19}\\ & zipper & 99.31\std{0.15} & 99.63\std{0.25} & 98.52\std{0.32} & 95.29\std{0.02} & 53.32\std{0.95} & 91.26\std{0.48}\\ \cmidrule(lr){2-8} & \textbf{Mean} & \textbf{97.06}\std{0.15} & \textbf{98.39}\std{0.19} & \textbf{96.21}\std{0.07} & \textbf{95.87}\std{0.02} & \textbf{56.19}\std{0.48} & \textbf{92.63}\std{0.03}\\
\midrule
\multirow{16}{*}{2-shot} & bottle & 99.92\std{0.11} & 99.97\std{0.04} & 99.81\std{0.28} & 99.05\std{0.37} & 78.06\std{0.77} & 96.57\std{0.47}\\
& cable & 97.33\std{1.67} & 98.08\std{1.08} & 93.22\std{2.39} & 93.80\std{2.72} & 58.89\std{1.70} & 85.75\std{3.08}\\
& capsule & 95.61\std{0.17} & 99.27\std{0.36} & 96.22\std{0.91} & 96.17\std{0.34} & 45.89\std{0.35} & 95.62\std{0.71}\\
& carpet & 99.84\std{0.17} & 99.95\std{0.05} & 99.64\std{0.08} & 99.15\std{0.12} & 61.19\std{0.18} & 97.44\std{0.45}\\
& grid & 100.00\std{0.00} & 100.00\std{0.00} & 100.00\std{0.00} & 99.20\std{0.01} & 42.55\std{0.18} & 96.50\std{0.04}\\
& hazelnut & 99.98\std{0.03} & 99.99\std{0.01} & 99.70\std{0.43} & 99.20\std{0.03} & 71.77\std{0.71} & 97.25\std{0.08}\\
& leather & 100.00\std{0.00} & 100.00\std{0.00} & 100.00\std{0.00} & 99.56\std{0.59} & 43.05\std{0.57} & 98.71\std{0.08}\\
& metal\_nut & 99.80\std{0.07} & 99.95\std{0.05} & 99.20\std{0.37} & 96.43\std{0.18} & 74.38\std{0.28} & 93.69\std{0.78}\\
& pill & 98.71\std{0.08} & 99.72\std{0.08} & 97.39\std{0.27} & 91.69\std{0.13} & 52.55\std{0.52} & 96.09\std{0.32}\\
& screw & 84.15\std{0.59} & 92.53\std{1.04} & 89.15\std{0.80} & 98.37\std{0.64} & 47.79\std{1.85} & 92.47\std{1.93}\\
& tile & 100.00\std{0.00} & 100.00\std{0.00} & 100.00\std{0.00} & 95.81\std{0.20} & 65.34\std{0.36} & 90.78\std{0.42}\\
& toothbrush & 93.99\std{2.11} & 98.00\std{0.66} & 94.38\std{0.77} & 96.87\std{0.93} & 35.00\std{4.05} & 93.46\std{1.38}\\
& transistor & 95.41\std{3.05} & 93.27\std{2.99} & 89.61\std{4.09} & 89.41\std{1.00} & 46.30\std{0.33} & 71.77\std{0.80}\\
& wood & 99.69\std{0.43} & 99.91\std{0.13} & 99.43\std{0.81} & 92.25\std{0.68} & 61.38\std{0.02} & 94.17\std{0.33}\\
& zipper & 98.91\std{0.01} & 99.68\std{0.01} & 98.64\std{0.15} & 96.32\std{0.49} & 54.58\std{0.23} & 92.07\std{0.47}\\
\cmidrule(lr){2-8}
& \textbf{Mean} & \textbf{97.56}\std{0.22} & \textbf{98.69}\std{0.09} & \textbf{97.09}\std{0.09} & \textbf{96.22}\std{0.08} & \textbf{55.91}\std{0.03} & \textbf{92.82}\std{0.07}\\
\midrule
\multirow{16}{*}{4-shot} & bottle & 100.00\std{0.00} & 100.00\std{0.00} & 100.00\std{0.00} & 98.95\std{0.05} & 80.64\std{2.08} & 96.86\std{0.20}\\
& cable & 96.18\std{0.97} & 97.82\std{0.55} & 92.92\std{0.92} & 94.09\std{0.73} & 61.25\std{1.88} & 87.54\std{0.67}\\
& capsule & 95.02\std{1.10} & 98.84\std{0.29} & 96.22\std{0.50} & 96.91\std{0.23} & 46.28\std{0.54} & 95.94\std{0.58}\\
& carpet & 99.80\std{0.06} & 99.96\std{0.03} & 99.44\std{0.11} & 99.06\std{0.00} & 61.65\std{1.07} & 97.62\std{0.23}\\
& grid & 100.00\std{0.00} & 100.00\std{0.00} & 100.00\std{0.00} & 99.22\std{0.01} & 42.72\std{0.59} & 96.52\std{0.31}\\
& hazelnut & 100.00\std{0.00} & 100.00\std{0.00} & 100.00\std{0.00} & 99.26\std{0.51} & 73.04\std{0.59} & 97.54\std{0.24}\\
& leather & 100.00\std{0.00} & 100.00\std{0.00} & 100.00\std{0.00} & 99.07\std{0.37} & 44.59\std{1.33} & 98.64\std{0.05}\\
& metal\_nut & 99.74\std{0.14} & 99.94\std{0.03} & 99.10\std{0.32} & 97.08\std{0.18} & 77.91\std{0.97} & 94.09\std{0.33}\\
& pill & 97.74\std{0.76} & 99.61\std{0.15} & 97.05\std{0.43} & 92.20\std{0.25} & 53.80\std{0.62} & 96.45\std{0.05}\\
& screw & 90.28\std{2.80} & 96.04\std{2.07} & 91.29\std{0.12} & 98.69\std{0.42} & 55.69\std{2.33} & 93.84\std{0.36}\\
& tile & 100.00\std{0.00} & 100.00\std{0.00} & 100.00\std{0.00} & 95.90\std{0.22} & 65.93\std{0.53} & 91.16\std{0.18}\\
& toothbrush & 97.03\std{2.25} & 98.97\std{0.57} & 96.68\std{2.90} & 97.43\std{0.58} & 38.73\std{2.66} & 94.67\std{1.07}\\
& transistor & 97.90\std{0.10} & 96.38\std{0.77} & 94.49\std{0.65} & 90.79\std{0.10} & 48.31\std{0.57} & 73.11\std{0.54}\\
& wood & 99.65\std{0.31} & 99.89\std{0.10} & 98.91\std{0.95} & 91.94\std{0.06} & 61.99\std{0.63} & 94.07\std{0.03}\\
& zipper & 99.15\std{0.10} & 99.74\std{0.03} & 99.30\std{0.48} & 96.37\std{0.18} & 55.52\std{0.81} & 92.59\std{0.21}\\
\cmidrule(lr){2-8}
& \textbf{Mean} & \textbf{98.17}\std{0.15} & \textbf{99.15}\std{0.16} & \textbf{97.69}\std{0.02} & \textbf{96.47}\std{0.07} & \textbf{57.87}\std{0.72} & \textbf{93.38}\std{0.01}\\
\end{longtable}
}

{\scriptsize
\captionsetup{font=footnotesize}
\setlength{\tabcolsep}{2pt}
\begin{longtable}{@{}c l ccc ccc@{}}
\caption{Per-category results on VisA under the 1-, 2-, and 4-shot settings, averaged over three random seeds (mean$\pm$std).}\label{tab:supp_visa_percat}\\
\toprule
& & \multicolumn{3}{c}{Image-level} & \multicolumn{3}{c}{Pixel-level}\\
\cmidrule(lr){3-5}\cmidrule(lr){6-8}
& Category & AUROC & AP & F1 & AUROC & F1 & PRO\\
\midrule
\endfirsthead
\multicolumn{8}{c}{\tablename~\thetable\ -- continued from previous page}\\
\toprule
& & \multicolumn{3}{c}{Image-level} & \multicolumn{3}{c}{Pixel-level}\\
\cmidrule(lr){3-5}\cmidrule(lr){6-8}
& Category & AUROC & AP & F1 & AUROC & F1 & PRO\\
\midrule
\endhead
\bottomrule
\endfoot
\multirow{13}{*}{1-shot} & candle & 95.30\std{0.44} & 95.22\std{0.69} & 90.00\std{0.61} & 98.85\std{0.09} & 47.42\std{1.71} & 96.10\std{0.59}\\
& capsules & 91.72\std{1.91} & 95.54\std{1.14} & 88.72\std{3.27} & 97.14\std{0.60} & 42.23\std{0.11} & 94.59\std{0.88}\\
& cashew & 91.43\std{2.99} & 95.67\std{2.05} & 89.72\std{5.08} & 98.18\std{0.30} & 43.73\std{2.26} & 96.62\std{0.10}\\
& chewinggum & 99.20\std{0.10} & 99.59\std{0.03} & 96.59\std{0.31} & 99.54\std{0.10} & 63.72\std{7.07} & 95.38\std{0.74}\\
& fryum & 94.18\std{0.07} & 97.28\std{0.02} & 91.65\std{0.50} & 93.34\std{3.27} & 33.69\std{1.95} & 89.68\std{1.15}\\
& macaroni1 & 95.90\std{0.64} & 96.13\std{1.08} & 92.68\std{0.80} & 99.28\std{0.26} & 34.52\std{1.65} & 94.16\std{1.83}\\
& macaroni2 & 75.58\std{1.21} & 75.02\std{3.72} & 72.22\std{0.65} & 98.56\std{0.37} & 27.22\std{1.56} & 90.31\std{2.10}\\
& pcb1 & 91.84\std{1.96} & 89.53\std{2.11} & 85.67\std{1.61} & 99.09\std{0.03} & 65.29\std{3.83} & 90.34\std{1.34}\\
& pcb2 & 88.72\std{0.87} & 85.88\std{0.79} & 82.95\std{0.57} & 96.48\std{0.35} & 40.44\std{1.62} & 81.76\std{2.50}\\
& pcb3 & 86.56\std{4.58} & 89.15\std{4.39} & 81.35\std{5.25} & 96.95\std{0.55} & 29.79\std{1.66} & 83.55\std{0.69}\\
& pcb4 & 90.60\std{3.95} & 91.42\std{3.53} & 85.11\std{5.95} & 93.40\std{1.41} & 28.63\std{5.96} & 68.61\std{3.32}\\
& pipe\_fryum & 96.82\std{0.24} & 98.33\std{0.21} & 95.63\std{0.79} & 98.24\std{0.49} & 41.83\std{1.01} & 97.43\std{0.52}\\
\cmidrule(lr){2-8}
& \textbf{Mean} & \textbf{91.49}\std{0.18} & \textbf{92.40}\std{0.10} & \textbf{87.69}\std{0.55} & \textbf{97.42}\std{0.16} & \textbf{41.54}\std{0.81} & \textbf{89.88}\std{0.92}\\
\midrule
\multirow{13}{*}{2-shot} & candle & 96.38\std{0.21} & 96.43\std{0.16} & 90.64\std{0.25} & 99.15\std{0.06} & 47.27\std{1.77} & 95.68\std{0.35}\\
& capsules & 92.86\std{0.30} & 96.21\std{0.39} & 88.96\std{0.88} & 97.22\std{0.39} & 48.64\std{2.16} & 92.95\std{1.61}\\
& cashew & 95.24\std{0.20} & 97.41\std{0.45} & 92.16\std{1.15} & 98.18\std{0.23} & 44.25\std{1.36} & 96.20\std{1.64}\\
& chewinggum & 98.85\std{0.31} & 99.34\std{0.21} & 97.30\std{0.20} & 99.65\std{0.03} & 66.43\std{1.30} & 95.90\std{0.29}\\
& fryum & 92.48\std{1.50} & 96.65\std{0.58} & 89.16\std{2.35} & 93.63\std{0.05} & 35.14\std{0.70} & 89.59\std{0.52}\\
& macaroni1 & 92.47\std{0.88} & 92.29\std{0.88} & 88.45\std{0.94} & 99.58\std{0.02} & 34.57\std{1.43} & 93.77\std{0.39}\\
& macaroni2 & 78.27\std{1.06} & 78.75\std{1.34} & 72.83\std{1.97} & 98.59\std{0.30} & 26.51\std{0.92} & 91.00\std{1.55}\\
& pcb1 & 93.33\std{0.41} & 89.78\std{0.49} & 87.44\std{0.25} & 99.15\std{0.10} & 63.92\std{1.54} & 90.97\std{0.45}\\
& pcb2 & 91.33\std{0.51} & 87.31\std{0.23} & 86.91\std{1.48} & 97.36\std{0.33} & 42.05\std{0.94} & 83.01\std{0.61}\\
& pcb3 & 89.82\std{1.21} & 91.25\std{0.13} & 82.63\std{0.40} & 96.99\std{0.08} & 28.42\std{1.10} & 83.75\std{0.69}\\
& pcb4 & 97.72\std{0.21} & 97.09\std{0.11} & 95.74\std{0.71} & 95.63\std{0.42} & 26.98\std{0.70} & 79.81\std{1.80}\\
& pipe\_fryum & 98.21\std{0.24} & 98.57\std{0.67} & 96.99\std{0.86} & 98.16\std{0.05} & 41.68\std{0.68} & 96.90\std{0.54}\\
\cmidrule(lr){2-8}
& \textbf{Mean} & \textbf{93.08}\std{0.59} & \textbf{93.42}\std{0.22} & \textbf{89.10}\std{0.14} & \textbf{97.77}\std{0.02} & \textbf{42.15}\std{0.50} & \textbf{90.79}\std{0.24}\\
\midrule
\multirow{13}{*}{4-shot} & candle & 96.56\std{1.11} & 96.73\std{1.39} & 90.48\std{0.48} & 99.04\std{0.08} & 47.72\std{0.11} & 96.15\std{0.54}\\
& capsules & 93.47\std{0.56} & 96.43\std{0.52} & 89.80\std{0.49} & 97.47\std{0.51} & 50.67\std{0.71} & 94.94\std{0.48}\\
& cashew & 95.37\std{0.81} & 96.78\std{0.30} & 90.67\std{0.49} & 98.72\std{0.04} & 46.91\std{0.23} & 96.82\std{0.35}\\
& chewinggum & 99.09\std{0.13} & 99.08\std{0.66} & 96.73\std{0.30} & 99.58\std{0.09} & 65.39\std{5.39} & 95.75\std{0.12}\\
& fryum & 94.90\std{0.17} & 97.70\std{0.05} & 91.36\std{0.11} & 94.50\std{0.28} & 37.47\std{0.99} & 91.31\std{0.07}\\
& macaroni1 & 93.81\std{1.38} & 94.25\std{1.05} & 87.44\std{2.81} & 99.61\std{0.06} & 34.06\std{1.15} & 94.37\std{0.49}\\
& macaroni2 & 78.66\std{2.41} & 79.98\std{1.83} & 72.88\std{1.96} & 98.72\std{0.32} & 24.34\std{1.22} & 90.47\std{0.73}\\
& pcb1 & 94.08\std{0.41} & 91.30\std{0.21} & 89.28\std{0.08} & 99.21\std{0.35} & 64.84\std{2.23} & 91.41\std{0.67}\\
& pcb2 & 91.59\std{0.93} & 87.50\std{0.83} & 86.07\std{0.31} & 97.58\std{0.10} & 42.47\std{1.31} & 84.21\std{1.45}\\
& pcb3 & 93.14\std{1.44} & 94.39\std{0.73} & 85.64\std{0.10} & 97.61\std{0.54} & 31.41\std{0.26} & 85.21\std{0.98}\\
& pcb4 & 98.70\std{0.61} & 98.25\std{0.19} & 97.08\std{1.36} & 96.40\std{0.38} & 30.98\std{0.47} & 84.03\std{1.94}\\
& pipe\_fryum & 97.80\std{0.36} & 98.76\std{0.06} & 96.47\std{0.36} & 98.25\std{0.18} & 40.72\std{0.78} & 97.41\std{0.42}\\
\cmidrule(lr){2-8}
& \textbf{Mean} & \textbf{93.93}\std{0.31} & \textbf{94.26}\std{0.01} & \textbf{89.49}\std{0.24} & \textbf{98.06}\std{0.01} & \textbf{43.08}\std{0.45} & \textbf{91.84}\std{0.09}\\
\end{longtable}
}

\begin{table}[htbp]
\centering
\caption{Cross-domain results on BraTS under the 1-, 2-, and 4-shot settings, averaged over three random seeds (mean$\pm$std). BraTS is single-class in the BMAD reorganization~\cite{bao2024bmad}, so no per-category breakdown applies.}
\label{tab:supp_brats_percat}
\begin{tabular}{l ccc ccc}
\toprule
& \multicolumn{3}{c}{Image-level} & \multicolumn{3}{c}{Pixel-level}\\
\cmidrule(lr){2-4}\cmidrule(lr){5-7}
Setting & AUROC & AP & F1 & AUROC & F1 & PRO\\
\midrule
1-shot & 73.69\std{4.14} & 92.12\std{1.43} & 91.55\std{0.79} & 95.81\std{0.84} & 46.44\std{4.83} & 76.14\std{3.64}\\
2-shot & 79.83\std{4.16} & 94.04\std{1.33} & 92.10\std{0.64} & 96.53\std{0.53} & 51.06\std{4.03} & 78.50\std{1.61}\\
4-shot & 83.80\std{2.59} & 95.39\std{0.83} & 92.38\std{0.59} & 96.97\std{0.29} & 54.32\std{2.39} & 79.76\std{1.26}\\
\bottomrule
\end{tabular}
\end{table}

\section{Additional Ablations}
\label{supp:ablation}

Table~\ref{tab:supp_controls} reports control experiments on VisA~$\rightarrow$~MVTec AD
under the 1-shot setting, averaged over three seeds. All trained rows share the same
backbone and differ from (g) only in the stated component; the rows are not cumulative.

\begin{table}[htbp]
\centering
\caption{Control experiments (VisA~$\rightarrow$~MVTec AD, 1-shot, three seeds). Rows are not cumulative.}
\label{tab:supp_controls}
\small
\setlength{\tabcolsep}{4pt}
\begin{tabular}{l ccc ccc}
\toprule
& \multicolumn{3}{c}{Image-level} & \multicolumn{3}{c}{Pixel-level}\\
\cmidrule(lr){2-4}\cmidrule(lr){5-7}
Configuration & AUROC & AP & F1 & AUROC & F1 & PRO\\
\midrule
$S_{\mathrm{image}}$ only (training-free) & 93.79 & 96.60 & 94.77 & 95.61 & 54.21 & 91.48\\
\midrule
(a) gate $\equiv 1$ & 93.99 & 96.58 & 94.61 & 95.03 & 54.91 & 92.22\\
(b) single direction $(\bar{t}^{a}-\bar{t}^{n})$ & 93.93 & 96.60 & 94.45 & 95.44 & 54.18 & 92.04\\
(c) generic prompts & 95.09 & 97.56 & 95.86 & 95.64 & 54.82 & 91.69\\
(d) wrong-dataset taxonomy & 95.32 & 97.45 & 95.47 & 95.81 & 55.09 & 92.15\\
(e) mean-centering & 97.05 & 98.39 & 96.17 & 95.87 & 56.19 & 92.63\\
(f) w/o $\mathcal{L}_{\mathrm{align}}$ & 95.52 & 97.37 & 95.19 & 95.74 & 55.08 & 92.26\\
\midrule
(g) ours & \textbf{97.06} & \textbf{98.39} & \textbf{96.21} & \textbf{95.87} & \textbf{56.19} & \textbf{92.63}\\
\bottomrule
\end{tabular}
\end{table}

\subsubsection{Where the gain comes from.}
Row (a) is a capacity-matched control: identical $g_I$, $h_S$ and identical losses,
including $\mathcal{L}_{\mathrm{align}}$, with the gate fixed to $g_i\equiv 1$. It accounts
for only a small part of the gain over $S_{\mathrm{image}}$, so the improvement is
attributable to the gate rather than to added trainable capacity or mask supervision.
Row (f) shows that $\mathcal{L}_{\mathrm{align}}$ is not an independent route to
discriminative patch features: without the gate the model barely moves, consistent with
its role as a precondition for the patch-wise matching.

\subsubsection{What the gate exploits.}
Row (b) collapses the candidate set to $\{\bar{t}^{n},\bar{t}^{a}\}$ and falls below the
gateless control, so a fixed single direction contributes nothing; the gain lies in soft
matching over many candidates. Row (e) replaces the normal anchor with the mean of all
text candidates and matches (g), indicating that subtracting a fixed reference direction
is what matters rather than the identity of the anchor. That (d) is not below (c) further
shows that incorrect defect terms do not mislead the gate: what it exploits is the spread
of the candidate set away from the anchor, not the correctness of any single term.

\subsubsection{Robustness to generic prompts.}
Row (c) restricts the abnormal prompts to generic descriptors such as ``damaged'',
``defective'' and ``anomalous''. A substantial part of the gain survives without any
fine-grained defect vocabulary, although the margin shrinks. The same holds in the
cross-domain setting: on MVTec AD~$\rightarrow$~BraTS with generic prompts and the terms
``tumor'', ``lesion'' and ``mass'' removed, TIRAD reaches $73.03/78.84/82.19$ image-level
AUROC at $1$-/$2$-/$4$-shot, against $72.79/76.96/80.53$ for the best baseline; the
ordering holds at every shot count.

\section{Limitations and Broader Impact}
\label{supp:limit}

\subsubsection{Structural limitation of multiplicative gating.}
The text branch modulates the image residual multiplicatively,
$\bar{r}^{I}_i = g_i \odot \tilde{r}^{I}_i$ with $g_i\in(0,1)$. Consequently the gate can
only attenuate or preserve spatial signal already present in the image residual; it
cannot introduce response in regions where the image residual is near zero. Localization
gains from the text branch are therefore bounded by the spatial support of the image
branch, and the method is best understood as semantic refinement of visual residuals
rather than an independent localizer. This is consistent with our finding that the
clearest gains appear at the image level and in the cross-domain regime.

\subsubsection{Prompt dependence.}
The size of the gain depends on the abnormal prompt set. With generic descriptors alone
the ordering against the baselines is preserved but the margins shrink
(Sec.~\ref{supp:ablation}), so reducing this dependence, for example by generating
candidates without a predefined taxonomy, remains open. We also note that the control
experiments in Sec.~\ref{supp:ablation} are reported as three-seed means without paired
uncertainty estimates.

\subsubsection{Mixed patches.}
As noted in the main paper, a patch containing both normal and abnormal regions receives
an ambiguous label between the normal and abnormal textual states, weakening the text
residual and blurring boundary localization; finer-grained or soft patch-level
supervision is a natural direction.

\subsubsection{Single medical benchmark.}
Our cross-domain evidence rests on one medical dataset (BraTS). Broader validation across
additional medical modalities and other domain shifts is needed to establish the
generality of the semantic-prior claim; we regard the current result as indicative rather
than conclusive.

\subsubsection{Broader impact.}
The medical prompts are generic semantic descriptors used only to form the normal and abnormal text features; they are not diagnostic statements, and the method is a research prototype not
intended for clinical use. All datasets used are publicly available and contain no
personally identifying information.

\section{Additional Qualitative Results}
\label{supp:qual}

\begin{figure}[htbp]
\centering
\includegraphics[width=1.0\linewidth,trim={50pt 100pt 50pt 100pt}, clip]{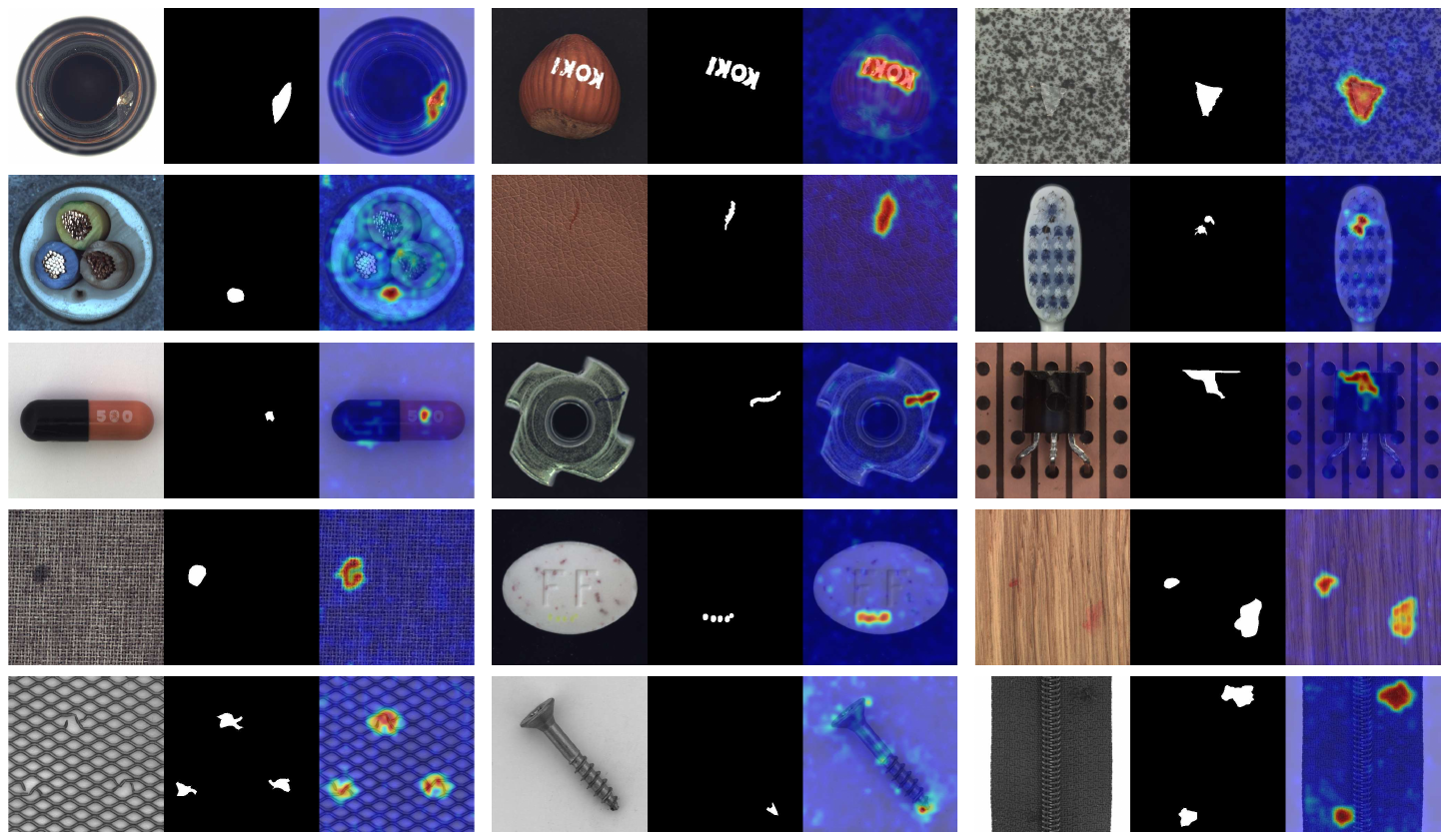}
\caption{Additional qualitative results on MVTec AD.}
\label{fig:supp_mvtec}
\end{figure}

\begin{figure}[htbp]
\centering
\includegraphics[width=1.0\linewidth,trim={50pt 120pt 50pt 120pt}, clip]{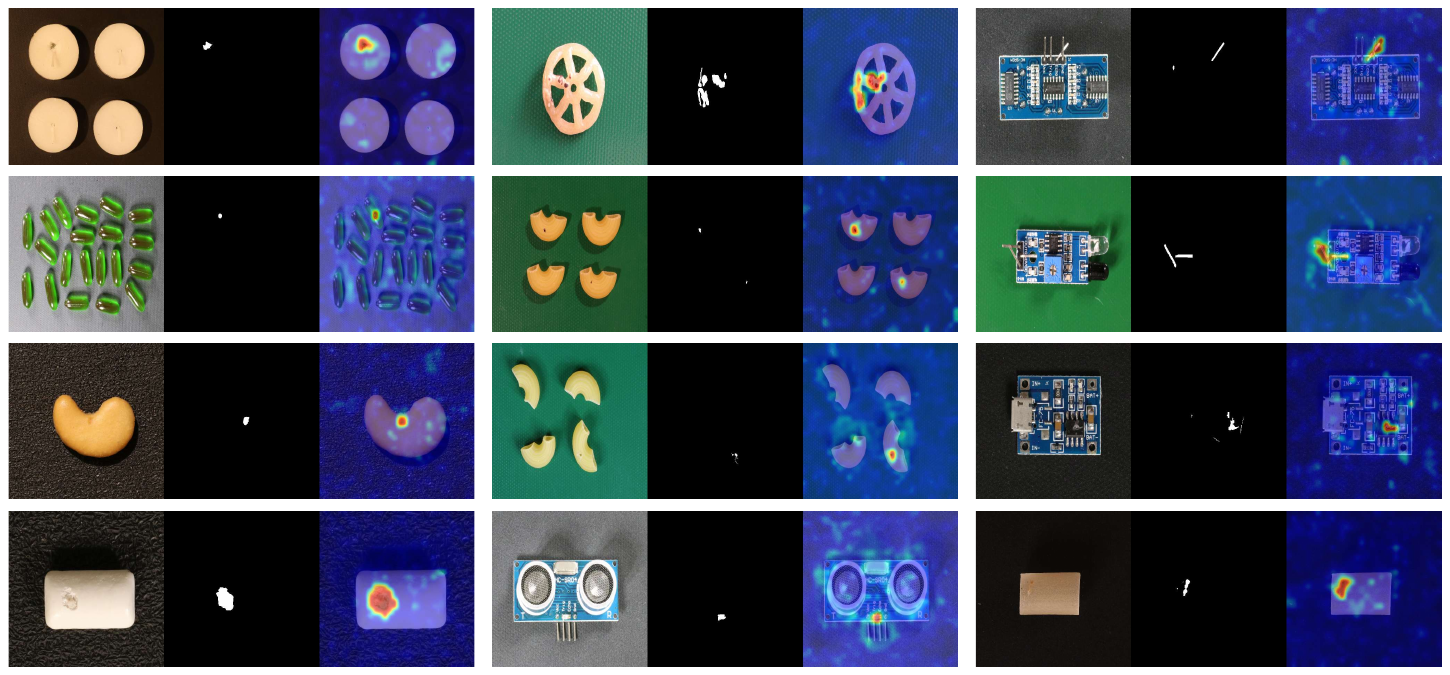}
\caption{Additional qualitative results on VisA.}
\label{fig:supp_visa}
\end{figure}

\begin{figure}[htbp]
\centering
\includegraphics[width=1.0\linewidth,trim={50pt 180pt 50pt 180pt}, clip]{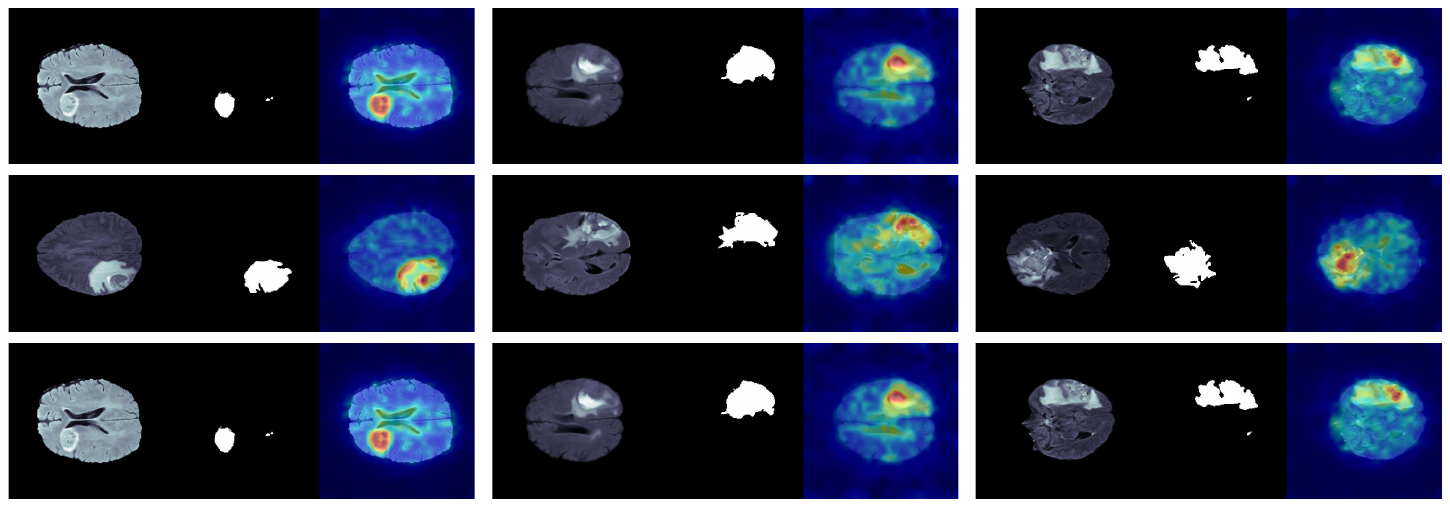}
\caption{Additional qualitative results on BraTS.}
\label{fig:supp_brain}
\end{figure}

Figures~\ref{fig:supp_mvtec}, \ref{fig:supp_visa}, and~\ref{fig:supp_brain} show
additional qualitative results on MVTec AD, VisA, and BraTS. For each example, from left
to right, we show the input image, the ground-truth anomaly mask, and the predicted
anomaly map.

\end{document}